\documentclass[pdflatex,sn-mathphys-num]{sn-jnl}

\usepackage[T1,OT1]{fontenc}
\DeclareTextAccentDefault{\k}{T1}
\usepackage{graphicx}%
\usepackage{array}
\usepackage{multirow}%
\usepackage{amsmath,amssymb,amsfonts}%
\usepackage{amsthm}%
\usepackage{mathrsfs}%
\usepackage[title]{appendix}%
\usepackage{xcolor}%
\usepackage[normalem]{ulem}%
\usepackage{textcomp}%
\usepackage{manyfoot}%
\usepackage{booktabs}%
\usepackage{algorithm}%
\usepackage{algorithmicx}%
\usepackage{algpseudocode}%
\usepackage{listings}%
\usepackage[nolist,nohyperlinks]{acronym}
\usepackage{array}
\usepackage{cleveref}
\usepackage{subcaption}
\usepackage{rotating}
\usepackage{xspace}
\usepackage{lineno}
\usepackage{verbatim}
\usepackage{tabularx}
\usepackage{longtable,booktabs}
\usepackage{tikz}
\usepackage{pdflscape}
\usetikzlibrary{arrows.meta, positioning, shapes.geometric}

\theoremstyle{thmstyleone}%
\theoremstyle{thmstyletwo}%

\theoremstyle{thmstylethree}%

\newcommand{\graybrackets}[1]{\textsubscript{\textcolor{black}{#1}}}

\begin{document}

\begin{acronym}
    \acro{AI}{artificial intelligence}
    \acro{AUROC}{area under the receiver operating characteristic curve}
    \acro{BMI}{body mass index}
    \acro{CI}{confidence intervals}
    \acro{CXR}{chest X-ray}
    \acro{EHR}{electronic health record}
    \acro{FPN}{feature pyramid network}
    \acro{LLM}{large language model}
    \acro{MIM}{masked image modelling}
    \acro{MLLM}{multimodal large-language model}
    \acro{PHI}{protected health information}
    \acro{PTX}{pneumothorax}
    \acro{SSL}{self-supervised learning}
    \acro{SOTA}{state-of-the-art}
    \acro{ViT}{vision transformer} 
    \acro{VQA}{visual question answering}
    \acro{LT}[L\&T]{lines and tubes}
    \acro{MAIRA}{Multimodal AI Radiology Assistant}
    \acro{ICU}{intensive care unit}
    \acro{CVC}{central venous catheter}
    \acro{ETT}{endotracheal tube}
    \acro{PICC}{peripherally inserted central catheter}
    \acro{IABP}{intra-aortic balloon pump}
    \acro{NGT}{nasogastric tube}
    \acro{SGC}{Swan-Ganz catheter}
    \acro{IJCVC}[IJ CVC]{internal jugular central venous catheter}
\end{acronym}

\newcommand{\mayodataset}{CXR-MAYO-REPORT-GEN\xspace}
\newcommand{\mairatwo}{MAIRA-2\xspace}
\newcommand{\mairax}{MAIRA-X\xspace}
\newcommand{\ltmetric}{RAD-LT-EVAL\xspace}
\newcommand{\raddino}{R\textsc{ad}-DINO\xspace}
\newcommand{\raddinox}{R\textsc{ad}-DINO-X\xspace}
\newcommand{\mimiccxr}{MIMIC-CXR\xspace}
\newcommand{\fone}{F$_1$\xspace}

\newcommand{\reportsection}[1]{\textit{#1}}
\newcommand{\findings}{\reportsection{Findings}\xspace}
\newcommand{\impression}{\reportsection{Impression}\xspace}
\newcommand{\indication}{\reportsection{Indication}\xspace}
\newcommand{\technique}{\reportsection{Technique}\xspace}
\newcommand{\comparison}{\reportsection{Comparison}\xspace}
\newcommand{\frontal}{\reportsection{Frontal}\xspace}
\newcommand{\lateral}{\reportsection{Lateral}\xspace}
\newcommand{\prior}{\reportsection{Prior}\xspace}


\title[A Multimodal AI Assistant for Longitudinal Chest X-Ray Reporting with Lines and Tubes]{A Multimodal AI Assistant for Longitudinal Chest X-Ray Reporting with Lines and Tubes}

\author[1]{Harshita Sharma} 
\equalcont{%
  These authors contributed equally to this work.\par
  \textsuperscript{\textdaggerdbl}These authors equally supervised this work.
}

\author[2]{Maxwell C. Reynolds} 
\equalcont{%
  These authors contributed equally to this work.\par
  \textsuperscript{\textdaggerdbl}These authors equally supervised this work.
}

\author[1]{Valentina Salvatelli} 
\equalcont{%
  These authors contributed equally to this work.\par
  \textsuperscript{\textdaggerdbl}These authors equally supervised this work.
}

\author[2]{Anne-Marie G. Sykes} 

\author[2]{Kelly K. Horst} 

\author[1]{Anton Schwaighofer} 

\author[1]{Maximilian Ilse} 

\author[3]{Olesya Melnichenko} 

\author[1]{Sam Bond-Taylor} 

\author*[1]{Fernando P\'erez-Garc\'ia}\email{fperezgarcia@microsoft.com}

\author[2]{Vamshi K. Mugu} 

\author[2]{Alex Chan} 

\author[2]{Ceylan Colak} 

\author[2]{Shelby A. Swartz} 

\author[2]{Motassem B. Nashawaty} 

\author[2]{Austin J. Gonzalez } 

\author[2]{Heather A. Ouellette} 

\author[2]{B. Selnur Erdal} 

\author[2]{Beth A. Schueler} 

\author[1,4]{Maria T. Wetscherek} 

 \author[5]{Noel Codella} 

\author[6]{Mohit Jain} 

\author[1]{Shruthi Bannur} 

\author[1]{Kenza Bouzid} 

\author[1]{Daniel C. Castro} 

\author[1]{Stephanie Hyland} 

\author*[2]{Panagiotis Korfiatis\textsuperscript{\textdaggerdbl}}\email{korfiatis.panagiotis@mayo.edu}

\author*[2]{Ashish Khandelwal\textsuperscript{\textdaggerdbl}} \email{drashish83@gmail.com}

\author[1]{Javier Alvarez-Valle\textsuperscript{\textdaggerdbl}}

\affil[1]{
    \orgname{Microsoft},
    \orgaddress{
        \street{21 Station Road},
        \city{Cambridge},
        \postcode{CB1 2FB},
        \country{United Kingdom}
    }
}

\affil[2]{
    \orgname{Mayo Clinic},
    \orgaddress{
        \street{200 First Street SW},
        \city{Rochester},
        \state{MN},
        \postcode{55905},
        \country{USA}
    }
}

\affil[3]{
    \orgname{Microsoft},
    \orgaddress{
        \street{Microsoft Building 99, 14820 NE 36th Street},
        \city{Redmond},
        \state{WA},
        \postcode{98052},
        \country{USA}
    }
}

\affil[4]{
    \orgname{University of Cambridge and Cambridge University Hospitals NHS Foundation Trust},
    \orgaddress{
        \street{Hills Road},
        \city{Cambridge},
        \postcode{CB2 0QQ},
        \country{United Kingdom}
    }
}

\affil[5]{
    \orgname{Microsoft},
    \orgaddress{
        \street{1 Microsoft Way},
        \city{Redmond},
        \state{Washington},
        \postcode{98052},
        \country{United States}
    }
}

\affil[6]{
    \orgname{Microsoft},
    \orgaddress{
        \street{Vigyan, No. 9, Lavelle Road},
        \city{Bengaluru},
        \state{Karnataka},
        \postcode{560001},
        \country{India}
    }
}

\abstract{  
Radiologists face growing workloads, and writing chest X-ray reports is repetitive, particularly when tracking medical devices such as lines and tubes across serial exams. AI-generated draft reports could ease this burden if they are accurate and clinically evaluated. Here we show that \mairax, a multimodal AI model developed using 2,427,411 Mayo Clinic studies, generates chest X-ray reports that account for prior exams and cover clinical findings and lines and tubes, outperforming the state of the art on language quality, clinical correctness and device reporting across test datasets. We develop an evaluation framework for detailed assessment of lines and tubes reporting. In a blinded retrospective study of 600 studies reviewed by nine radiologists, AI-generated reports have slightly more critical errors than original reports (4.6\% vs. 3.0\%; $p = 0.006$), but a similar share of acceptable sentences (97.4\% vs. 97.7\%). These results suggest that \mairax can assist radiologists, particularly in high-volume settings. }

\keywords{Generative AI, user evaluation, chest X-rays, report generation, lines and tubes, clinical deployment}



\maketitle

\acresetall

\section{Introduction}\label{sec1}

Radiology imaging plays a pivotal role in modern healthcare, with approximately 4.2 billion diagnostic examinations conducted globally each year~\cite{mahesh2022patient}, a figure that continues to grow as technological advancements proliferate and healthcare demands increase. Beyond the growing patient volumes, radiologists confront challenges like expanding screening guidelines, increasingly complex cases, and demographic shifts due to aging populations. These pressures are further exacerbated by workforce shortages and fatigue 
among radiologists, with 49\% of professionals in the field reporting burnout~\cite{bailey2022understanding}. In this context, AI-assisted radiology report generation emerges as a promising solution by streamlining radiology workflows, while preserving accuracy and improving consistency of draft reports~\cite{10.1145/3613904.3642013,yu2023evaluating}.

\acused{AI}

Radiologists draft the \findings section as a detailed description of observations of the radiology images from the study in question. This section includes normal findings as well as any abnormalities, such as signs of disease, masses, fractures, and supporting devices, including their location, severity (for pathological findings), and changes from prior exams. Within this broad field of radiology report generation, \acp{CXR} represent a significant area of focus~\cite{li2025automatic,sloan2024automated}. Among all the \ac{CXR} interpretation tasks, lines (or catheters) and tubes are the second most common type of abnormal finding on the radiograph~\cite{yi2020computer}, and are recommended as the first element to inspect when reviewing a chest X-ray image~\cite{jones_chest_nodate}.

\acused{LT}

Different types of lines and tubes, collectively referred to as \ac{LT} in this paper, are inserted into the patient’s body to supply fluids, medication and nutrition, monitor body functions, and provide other treatments in clinical settings~\cite{godoy2012chest}. Chest X-rays provide a readily available first-line imaging assessment of positioning of \ac{LT}, and of complications following their insertion.  This emphasizes the need for timely image interpretation, especially in high-throughput clinical environments. For example, in \acp{ICU} and emergency departments, frequent and precise \ac{LT} reporting is crucial, as 
patients may have several \acp{LT} in place simultaneously
and differences between their appearances on longitudinal scans can be nuanced and complex. 
Hence, \ac{LT} reporting is particularly demanding and repetitive for radiologists, and can lead to significant cognitive workloads and fatigue due to high volumes and the need for prolonged attention. By reporting both clinical findings and \acp{LT}, \ac{AI} has the potential to enhance radiologists' efficiency by reducing their cognitive workload, 
thereby improving turnaround times and patient safety. 

Recent advancements in AI-driven radiology reporting have demonstrated promising results, particularly in the domain of \ac{CXR} 
report generation. These include generalist biomedical models encompassing multiple imaging modalities and applications~\cite{sellergren2025medgemmatechnicalreport,zhou2024generalist,yang2024advancing,tu2023generalistbiomedicalai}, and \ac{CXR}-specialist report generation models~\cite{ZambranoChaves2025, zhang2025libraleveragingtemporalimages,chen2024chexagent,bannur2024maira2groundedradiologyreport}, which have consistently been shown to surpass generalist AI models for this task.
Among the specialist models, the MAIRA family of \acp{MLLM} \cite{hyland2024maira1specialisedlargemultimodal,bannur2024maira2groundedradiologyreport,pmlr-v259-sharma25a, srivastav-etal-2024-maira} has recently emerged for automated chest X-ray reporting. Specifically, \mairatwo \cite{bannur2024maira2groundedradiologyreport}, a state-of-the-art multimodal generative AI model for \ac{CXR} report generation,
excels at generating the \findings section of radiology reports by incorporating contextual information, such as multiple report sections, prior images, prior reports, and leveraging multiple image views. The model has consistently outperformed other generative AI systems on public datasets like \mimiccxr \cite{sellergren2025medgemmatechnicalreport}, demonstrating its effectiveness in addressing core challenges in AI-assisted radiology reporting. 

Building upon these advancements, this paper introduces 
\mairax, a next-generation multimodal AI model designed for longitudinal chest X-ray reporting, encompassing both clinical findings and \acp{LT}. \mairax was trained on a large-scale, multi-site clinical dataset from Mayo Clinic. We optimized \mairax for detailed and accurate reporting of \acp{LT} along with the typical \ac{CXR} pathological findings. Specifically for \acp{LT}, \mairax seeks to describe instances of nine types of frequently used \acp{LT}, namely, \acp{CVC}~\cite{muhm1997malposition}, \acp{PICC}~\cite{johansson2013advantages}, \acp{NGT}~\cite{vadivelu2023evolving}, \acp{ETT}~\cite{haas2014endotracheal}, chest tubes~\cite{anderson2022comprehensive}, \acp{SGC}~\cite{chatterjee2009swan}, \acp{IABP}~\cite{webb2015management}, mediastinal drains~\cite{wallen2002mediastinal}, and tracheostomy tubes~\cite{schmidt2008tracheostomy}, along with their tip locations, side-specific details, and changes over time.

To assess the utility of our models, we adopt a nuanced approach to evaluation that goes beyond traditional metrics and embraces more comprehensive \ac{LT}-specific criteria. Prior work in report generation, including \mairatwo, has primarily focused on the evaluation of the model's clinical performance in terms of detection of common chest pathologies, as measured by 
CheXpert 
\cite{smit2020chexbert, irvin2019chexpert} and LLM-as-a-judge methods such as RadFact \cite{bannur2024maira2groundedradiologyreport}. \mairax surpasses these standard approaches by incorporating an 
\ac{LT}-specific evaluation framework to assess the detailed accuracy of \ac{LT} reporting. Quantitative evaluation of \mairax for lexical quality, clinical accuracy, and \ac{LT}-specific performance provides strong evidence of its superiority over state-of-the-art report generation methods. 

Because lexical metrics such as BLEU and ROUGE can score random or templated sentences highly without reflecting clinical accuracy~\cite{boag2020baselines,zhao2024x}, we evaluate \mairax through complementary signals: lexical and classifier-based metrics for comparability with prior work, LLM-based metrics (RadFact, \ltmetric) sensitive to clinical meaning, and, as the primary clinical signal, a 
retrospective user evaluation study involving  
nine radiologists with varying levels of experience. This user evaluation includes pathological and \ac{LT}-specific assessments, and provides critical insights into the capabilities of the \mairax model. 

\bigskip

\noindent The key contributions of this paper are as follows. 

\vspace{5pt}

\begin{enumerate}
\item 
\mairax for clinical \ac{CXR} report generation: 
We introduce \mairax, a multimodal AI model designed for longitudinal chest X-ray report generation, including relevant descriptions of clinical findings and \acp{LT}.
\begin{enumerate}

\item  
\mairax is a \ac{CXR}-specialized report generation model trained by leveraging \mayodataset, a large-scale, multi-site, de-identified clinical dataset of 2.4 million studies from Mayo Clinic. 

\item  We developed \ltmetric, an LLM-based evaluation framework that enables large-scale, fine-grained evaluation of \ac{LT}-specific performance in generative AI models for longitudinal \ac{CXR} report generation. 
\item \mairax was benchmarked against the public \mairatwo baseline across three holdout datasets, and additionally evaluated on the public \mimiccxr test split after continual training, in comparison with prior work such as MedGemma 4B~\cite{sellergren2025medgemmatechnicalreport}, \mairatwo~\cite{bannur2024maira2groundedradiologyreport}, and LIBRA~\cite{zhang2025libraleveragingtemporalimages} (which were trained primarily with \mimiccxr).

\end{enumerate}

\item 
User-centric evaluation study: To assess the clinical utility of \mairax, we conducted a retrospective user evaluation study on 600 cases reviewed by nine radiologists (three reviews per case) of varying experience levels across two cohorts: one that matches the clinical deployment distribution and another where rarer \acp{LT} and tip positions were upsampled.

\end{enumerate}

\begin{figure}[htbp]
    \centering
    \includegraphics[trim={0 250bp 0 0}, clip, width=\textwidth]{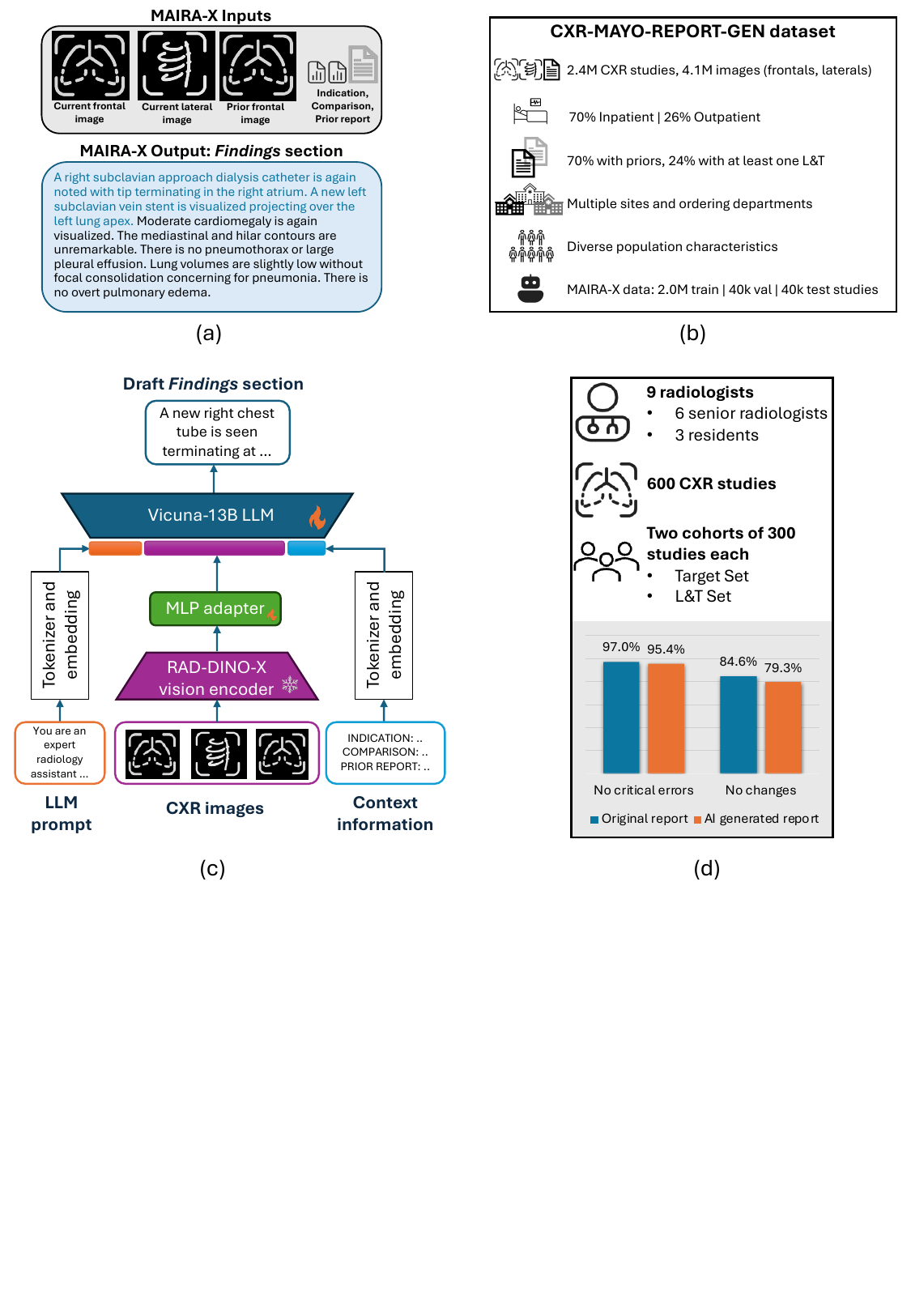}
    \caption{MAIRA-X overview. Overview of \mairax for longitudinal chest X-ray (\ac{CXR}) reporting including clinical findings and lines and tubes (\ac{LT}). (a) Problem definition (inputs and output shown for illustration only) (b) \mayodataset dataset: a large-scale, multi-site, clinical dataset from Mayo Clinic (c) \mairax model architecture (d) Summary of \mairax user evaluation study.}
\label{fig:maira_x_overview}
\end{figure}

\section{Results}\label{sec2}
We trained and evaluated \mairax on the \mayodataset dataset from Mayo Clinic, a large-scale, multi-site, longitudinal clinical dataset of 2,427,411 de-identified \ac{CXR} studies, comprising 4,123,776 images from 744,860 patients, acquired between 2007 and 2023. Details of the dataset, including patient demographics, 
are provided in \Cref{sec:dataset_details}.  
We optimized the model parameters by evaluating \mairax on a validation subset of \mayodataset (40,007 studies). Our quantitative evaluations were performed on multiple holdout subsets of \mayodataset with different sample sizes and distributions of \acp{LT}, including the test set (40,006 studies), 
the Target Set with \ac{LT} distribution mimicking the expected clinical setting (300 studies, 62 of which contain at least one \ac{LT}), and the \ac{LT} Set, a \ac{LT}-enriched (oversampled) cohort designed to stress-test performance on lines and tubes (300 studies, all with at least one \ac{LT}).
Details about splits and data subsets are provided in \Cref{sec:exp_setup}. We also report results on the \mimiccxr official test split for comparison with prior works in the literature. For the user evaluation study, we utilize 
the Target Set and \ac{LT} Set; details are provided in \Cref{sec:user_eval_setup}. An overview of the model inputs, \mayodataset, model architecture, and user evaluation study is shown in \Cref{fig:maira_x_overview}.

\subsection{\mairax significantly improves longitudinal chest X-ray reporting over state-of-the-art methods}


We evaluated \mairax quantitatively using lexical, clinical and \ac{LT}-specific metrics, and observed that \mairax outperforms the public \mairatwo baseline with remarkably improved performance on \acp{LT}. To evaluate the \ac{LT} performance, we developed an LLM-based structured report evaluation framework called \ltmetric. This framework was designed from the ground up with the radiologists,
and captures the key clinical aspects of \ac{LT} reporting such as their types, tip locations, changes from prior study, correctness of tip placements, and counts (see details of our evaluation metrics in \Cref{sec:quantitative_metrics}). 

\begin{figure}[htbp]
    \begin{center}
        \centering
        \includegraphics[width=\textwidth]{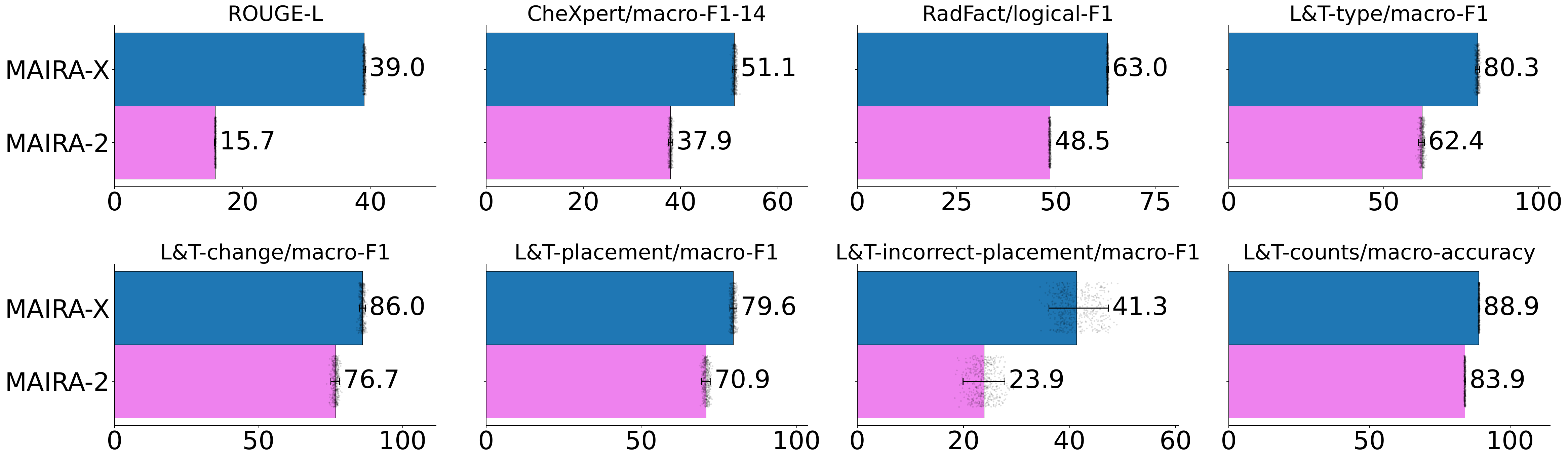}
    \caption{\mairax outperforms \mairatwo. Comparison of \mairax and \mairatwo on the \mayodataset holdout test set. Numeric values/center lines are median (ROUGE-L, CheXpert, RadFact) and mean (lines and tubes (\ac{LT}) metrics); error bars are 95\% \ac{CI} for $n=500$ bootstrapped samples. Raw bootstrap sample values are shown by small dots. Extended results, including those for other holdout datasets are presented in Supplementary Tables S4 and S5.}
    \label{fig:quantitative_results_1}
    \end{center}
\end{figure}

In~\Cref{fig:quantitative_results_1}, we report the lexical metric ROUGE-L~\cite{lin2004automatic}, clinical efficacy (CE) metrics, i.e., CheXpert/macro-\fone-14~\cite{smit2020chexbert} and RadFact/logical-\fone~\cite{bannur2024maira2groundedradiologyreport}, and \ltmetric metrics, i.e., \ac{LT}-type/macro-\fone (for detecting the \ac{LT} types), \ac{LT}-change/macro-\fone (for detecting the longitudinal change for each \ac{LT}), \ac{LT}-placement/macro-\fone (for detecting placement of each \ac{LT}) and \ac{LT}-counts/accuracy (for detecting the total number of reported \acp{LT}) on the holdout test set. We also report \ac{LT}-incorrect-placement/macro-\fone due to its clinical significance in \ac{CXR} reporting. We present detailed tables (Supplementary Tables S4 and S5) of results with additional metrics on the three holdout datasets. Per-pathology CheXpert results are reported in Supplementary Table S6 (\mairax compared with \mairatwo), Supplementary Table S7 (the full set of classification metrics for \mairax), and Supplementary Table S8 (stratified by presence of \acp{LT}). 

We observe that all quantitative metrics (i.e., lexical, clinical, and \ac{LT} structured report metrics of \ltmetric) consistently improve for \mairax compared to public \mairatwo~\cite{bannur2024maira2groundedradiologyreport}
on the three holdout sets. \mairax outperforms the public baseline by large margins, with 10 pp or more improvement on ROUGE-L, CheXpert/macro-\fone-14,  RadFact/logical-\fone, \ac{LT}-type/macro-\fone and \ac{LT}-incorrect-placement/macro-\fone.
Notably, these gains are observed despite MAIRA-2 being trained on multiple public datasets from different institutions.

Specifically for the \ac{LT}-specific metrics, we find that \mairax performs better in reporting \ac{LT} types, change from priors, and placements. For longitudinal change in particular, when evaluated across a five-class taxonomy (new, moved, removed, unchanged, unclear), \mairax achieves \ac{LT}-change/macro-\fone of 86.0 on the test set (vs.\ 76.7 for \mairatwo) and 87.5 on the validation set (vs.\ 78.4), with comparable gains on the micro-averaged variant. For incorrect \ac{LT} placement, the absolute \fone scores are comparatively low; 7.7\% of all \acp{LT} in the \mayodataset dataset are misplaced. Nevertheless, \mairax demonstrates significant improvements over the baseline. The \ac{LT} count accuracy is higher for \mairax, and the gap with \mairatwo increases as the number of \acp{LT} in the \ac{CXR} study increases from zero to three-or-more, with the largest gains (25 pp or more) for three or more \acp{LT}, suggesting its superior performance in critical settings such as the \ac{ICU}. Notably, the \ac{LT} Set, wherein reports have at least one line/tube, demonstrates the highest \ac{LT} type and placement metrics. Per-device \ac{LT} results for \mairax are reported in Supplementary Table S14 for the validation set and Supplementary Table S15 for the test set. 
The quantitative metrics show an overall performance gain for \mairax across the three spheres of text generation, namely, natural language generation (lexical metrics), clinical quality (clinical efficacy metrics), and line and tube reporting accuracy (\ac{LT} structured reporting metrics of \ltmetric).

\vfill
We quantitatively compare \mairax with existing work in radiology report generation using lexical and clinical performance metrics, namely ROUGE-L, CheXpert and RadFact. 
Specifically, we compare generalist models such as MedGemma 4B~\cite{sellergren2025medgemmatechnicalreport} and Med-PaLM M~\cite{tu2023generalistbiomedicalai} and \ac{CXR} specialist models like LLaVA-Rad~\cite{ZambranoChaves2025}, Libra~\cite{zhang2025libraleveragingtemporalimages}, and \mairatwo~\cite{bannur2024maira2groundedradiologyreport}.
We restrict this comparison to models that generate reports directly from imaging inputs; methods that rely on an intermediate class representation (e.g., predicted finding labels) before report generation are outside the scope of this benchmark.
For \ac{CXR} report generation, the existing models were predominantly trained on public datasets such as \mimiccxr and evaluated on the in-domain \mimiccxr test split. For a fair comparison, we continually trained the \mairax checkpoint on the \mimiccxr training split 
for one epoch and evaluated it on the official \mimiccxr test split. The results are demonstrated in~\Cref{tab:mairax_mimic}. For Med-PaLM M, LLaVA-Rad, Libra and \mairatwo, the metric values are directly reported from prior works. MedGemma 4B results were recomputed. \mairax outperforms the existing models for radiology report generation on the official \mimiccxr test split across the reported lexical and clinical metrics.

\subsection{Radiologists' assessments highlight the readiness of \mairax for deployment as a draft reporting tool}

\begin{figure}[b]  
    \centering  
    \includegraphics[trim={0cm, 9cm, 0cm, 0cm}, clip, width=\textwidth]{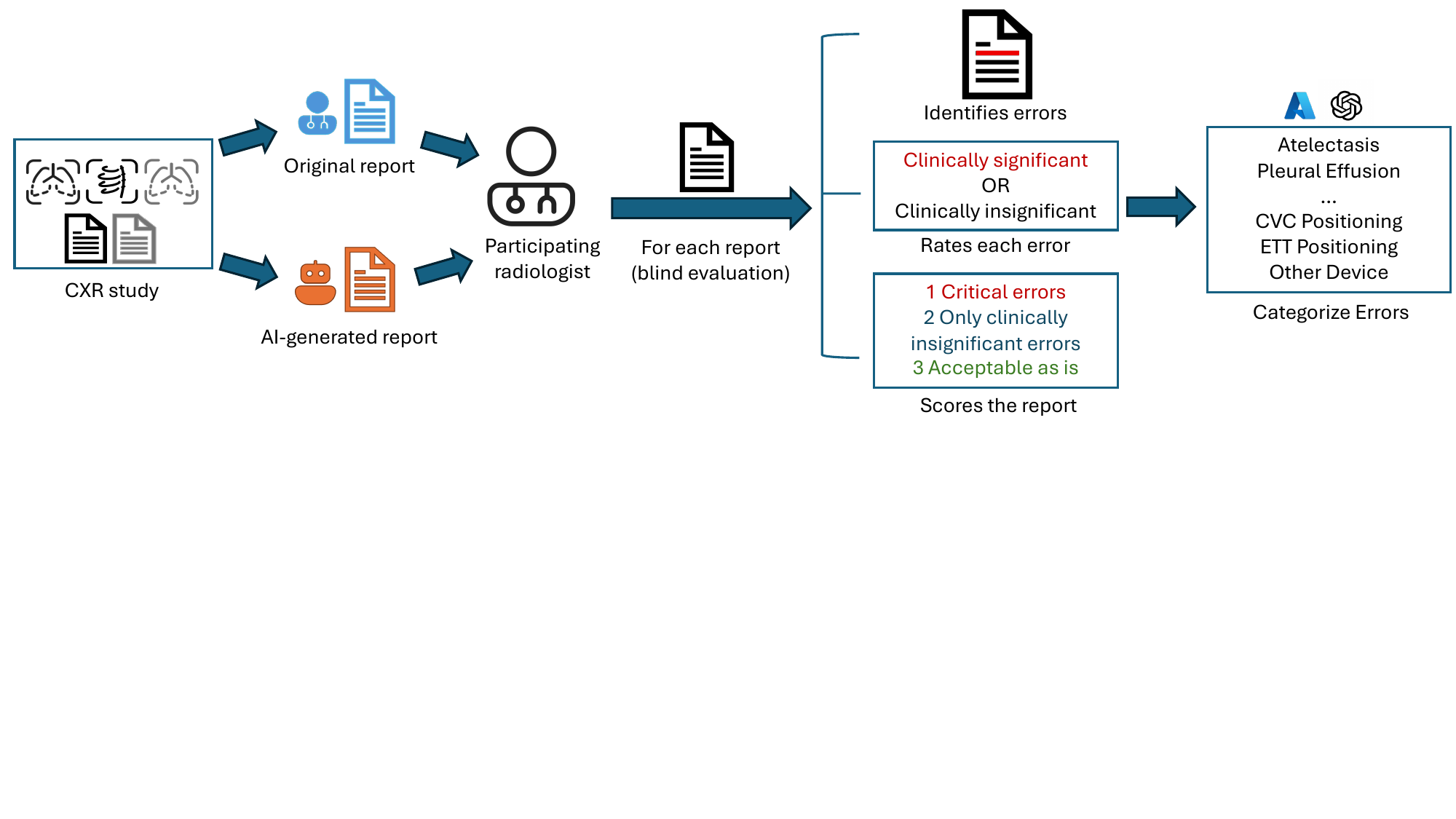}   
    \caption{Schematic overview of the \mairax user evaluation study. Radiologists see the original and AI-generated report from each chest X-ray (\ac{CXR}) study. Each report is rated by three radiologists in the user study, then errors are binned into categories using GPT-4o.} 
    \label{fig:user_eval_overview}  
\end{figure}

A schematic overview of the \mairax radiologists' evaluation study is depicted in \Cref{fig:user_eval_overview}. Of the 600 studies reviewed, 39 (6.5\%) were excluded because at least one of the three radiologist evaluations was missing or invalid, therefore, all results reported below are computed over 561 studies. Details of the user evaluation study, including the exclusion criteria, are provided in \Cref{sec:user_eval_setup}. Report quality in the radiologist user study was assessed using a predefined scoring rubric, where each report was assigned a score of 1 (clinically significant/critical errors), 2 (only clinically insignificant errors), or 3 (acceptable as-is). Clinically significant errors were defined as findings that may impact patient care or require communication with the clinician (e.g., missed important findings or incorrect line/tube placement), whereas clinically insignificant errors were defined as non-critical inaccuracies that do not affect patient management but require correction. From this rubric, we derive aggregate measures of report quality, including the proportion of reports with no critical errors and the proportion requiring no changes. Reviewers were instructed to disregard stylistic preferences from their review and focus solely on report content.

\begin{figure}[htbp]
    \centering
    \includegraphics[width=\textwidth]{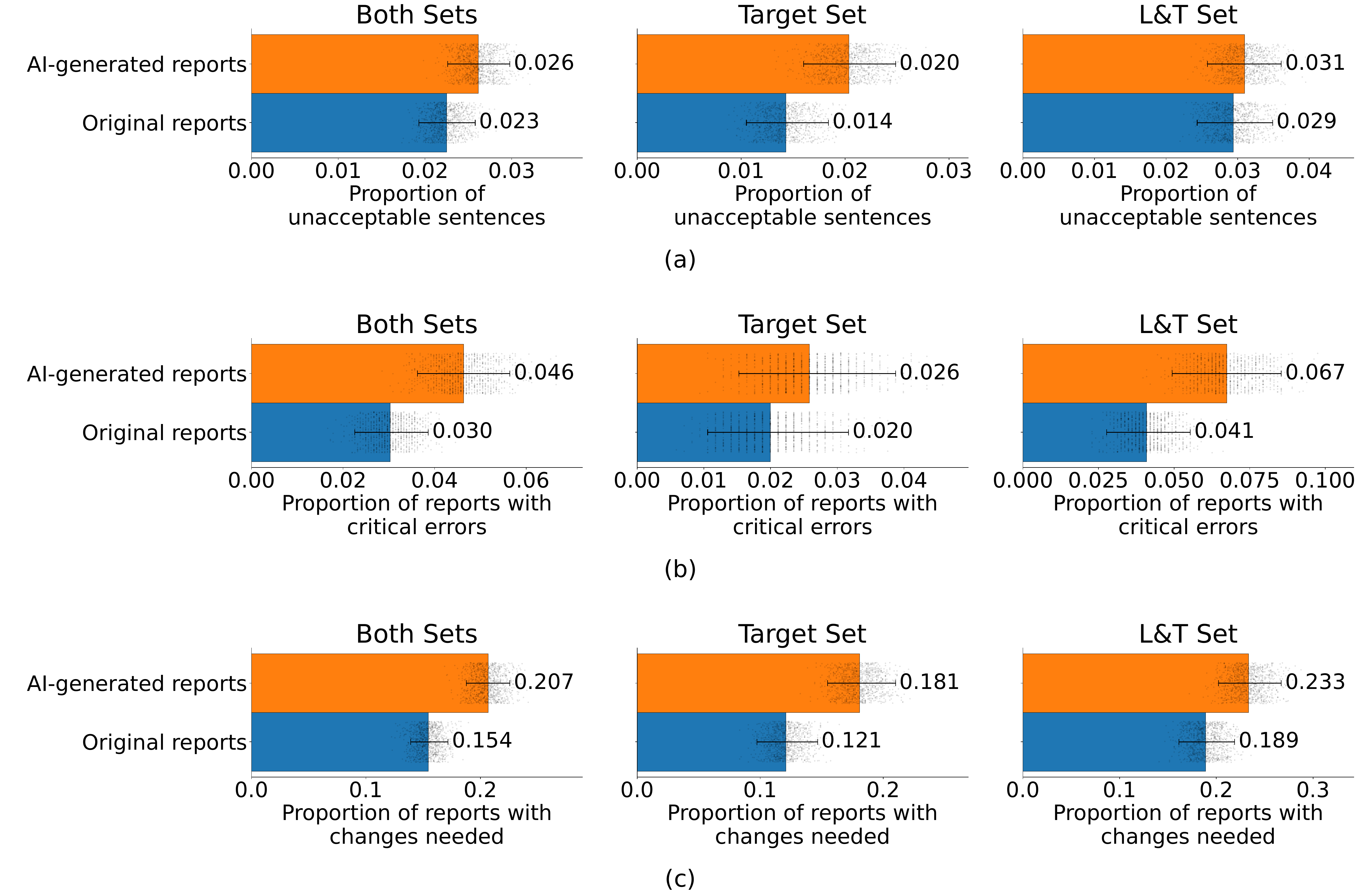}
    \caption{Radiologist-assessed error rates in original and MAIRA-X-generated reports. For original and \mairax-generated reports, proportions of (a) sentences requiring change (b) reports with critical errors and (c) reports with changes needed (i.e. any errors) are shown. Numeric values/center lines are the non-bootstrapped metrics. Error bars indicate 95\% confidence intervals obtained from 1,000 bootstrap resamples of the dataset. Raw bootstrap sample values are shown by small dots.}  
    \label{fig:comparison_plot}  
\end{figure}

The study results show similar report quality between original (i.e., radiologist-written) and AI-generated reports. As shown in \Cref{fig:comparison_plot}(a), the proportion of error-free sentences (no critical or clinically insignificant errors) is 97.7\% [$97.4\%-98.1\%$ \ac{CI}] in original reports and 97.4\% [$97.0\%-97.7\%$ \ac{CI}] in AI-generated reports. From \Cref{fig:comparison_plot}(b), we observe that, in aggregate (i.e., combining \ac{LT} and Target cohorts), 97.0\% [$96.1\%-97.7\%$ \ac{CI}] of original reports contain no critical errors, while 95.4\% [$94.4\%-96.4\%$ \ac{CI}] of AI-generated reports contain no critical errors. A one-sided permutation test (10,000 permutations) shows this difference to be statistically significant (difference in proportions, AI-generated minus original, $-1.60$ percentage points [$-2.85$ to $-0.36$, 95\% bootstrap \ac{CI}, 10,000 resamples]; $p = 0.006$). Moreover, \Cref{fig:comparison_plot}(c) demonstrates that 84.6\% [$82.8\%-86.1\%$ \ac{CI}] of original reports require no changes and are acceptable as-is, compared to 79.3\% [$77.4\%-81.2\%$ \ac{CI}] of AI-generated reports. A one-sided permutation test (10,000 permutations) also shows this difference to be statistically significant (difference in proportions, AI-generated minus original, $-5.23$ percentage points [$-7.55$ to $-2.85$, 95\% bootstrap \ac{CI}, 10,000 resamples]; $p < 0.001$).  A breakdown with respect to cohort (\ac{LT} and Target) is also shown in \Cref{fig:comparison_plot}. Performance in both original and AI-generated reports is stronger in the Target Set compared to the \ac{LT} Set.

\begin{figure}[htbp]
    \centering
    \includegraphics[width=\textwidth]{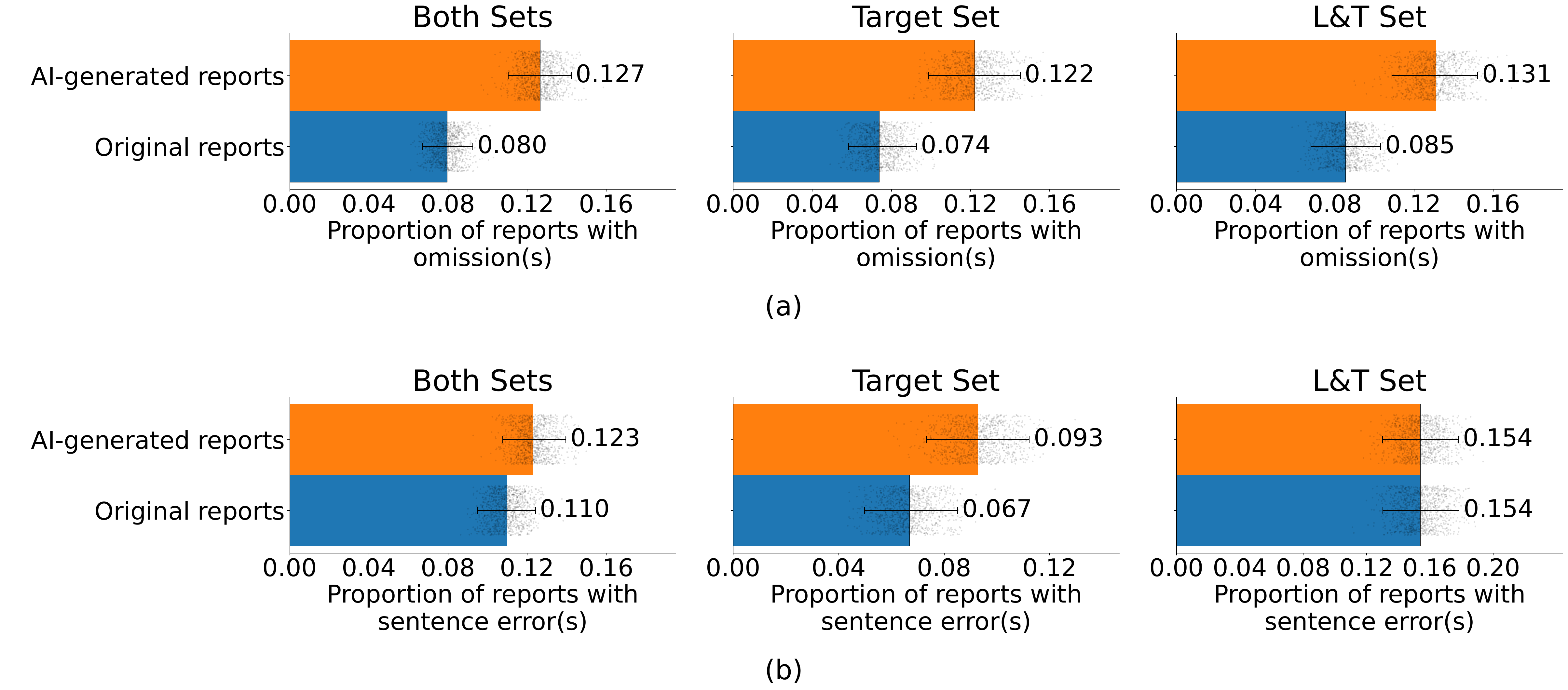}
    \caption{Omissions and sentence errors in original versus \mairax reports. Proportions of reports with (a) omissions and (b) sentence errors across original and \mairax-generated reports are shown. Numeric values/center lines are the non-bootstrapped metrics. Error bars indicate 95\% confidence intervals obtained from 1,000 bootstrap resamples of the dataset. Raw bootstrap sample values are shown by small dots.}  
    \label{fig:omissions_sentence_errors}  
\end{figure}

We define two possible error types. ``Omissions'' are defined as entire sentences (one sentence per finding or \ac{LT}) that are missing from the radiology report. ``Sentence Errors'' are defined as errors in the report that can be resolved by modifying an existing sentence. Possible examples of sentence-level error modifications include clarifying missing details of a finding or \ac{LT} (such as location or severity), removing hallucinations, changing a reported negative observation to a positive one, and correcting the interpreted underlying cause of a pathological finding. 
From \Cref{fig:omissions_sentence_errors}(a), we observe that 8.0\% [$6.7\%-9.3\%$ \ac{CI}] of original reports have at least one omission, while 12.7\% [$11.0\%-14.2\%$ \ac{CI}] of AI-generated reports have at least one omission.  Only 0.8\% [$0.4\%-1.2\%$ \ac{CI}] of original reports and 1.3\% [$0.8\%-1.9\%$ \ac{CI}] of AI-generated reports have multiple omissions. In \Cref{fig:omissions_sentence_errors}(b), we observe that sentence errors 
have similar proportions between original and AI-generated reports. Sentence errors are present in 11.0\% [$9.5\%-12.4\%$ \ac{CI}] of original reports and 12.3\% [$10.8\%-14.0\%$ \ac{CI}] of AI-generated reports. Only 1.8\% [$1.2\%-2.4\%$ \ac{CI}] of original reports and 1.7\% [$1.1\%-2.3\%$ \ac{CI}] of AI-generated reports have multiple sentence errors. Specifically for the \ac{LT} Set, the proportion of reports with sentence errors is nearly the same. Combining critical and non-critical errors, the mean number of errors per report was 0.22 [$0.19-0.24$ \ac{CI}] errors of any kind (0.03 [$0.03-0.04$ \ac{CI}] critical) for original reports, compared to 0.29 [$0.26-0.31$ \ac{CI}] errors of any kind (0.06 [$0.04-0.07$ \ac{CI}] critical) for \mairax-generated reports. We present qualitative examples of different errors (critical and non-critical; \ac{LT} and pathology) flagged across original and \mairax generated reports in \Cref{fig:qualitative1}, along with extended examples in Supplementary Figure S5. These include qualitative examples of both original and AI-generated reports that are either acceptable as-is, or contain errors (omissions, sentence errors) that could be either critical or clinically insignificant. 

\begin{sidewaysfigure}

    \centering

    \includegraphics[trim={0.5cm, 0cm, 0.5cm, 0cm}, clip, width=\textwidth]{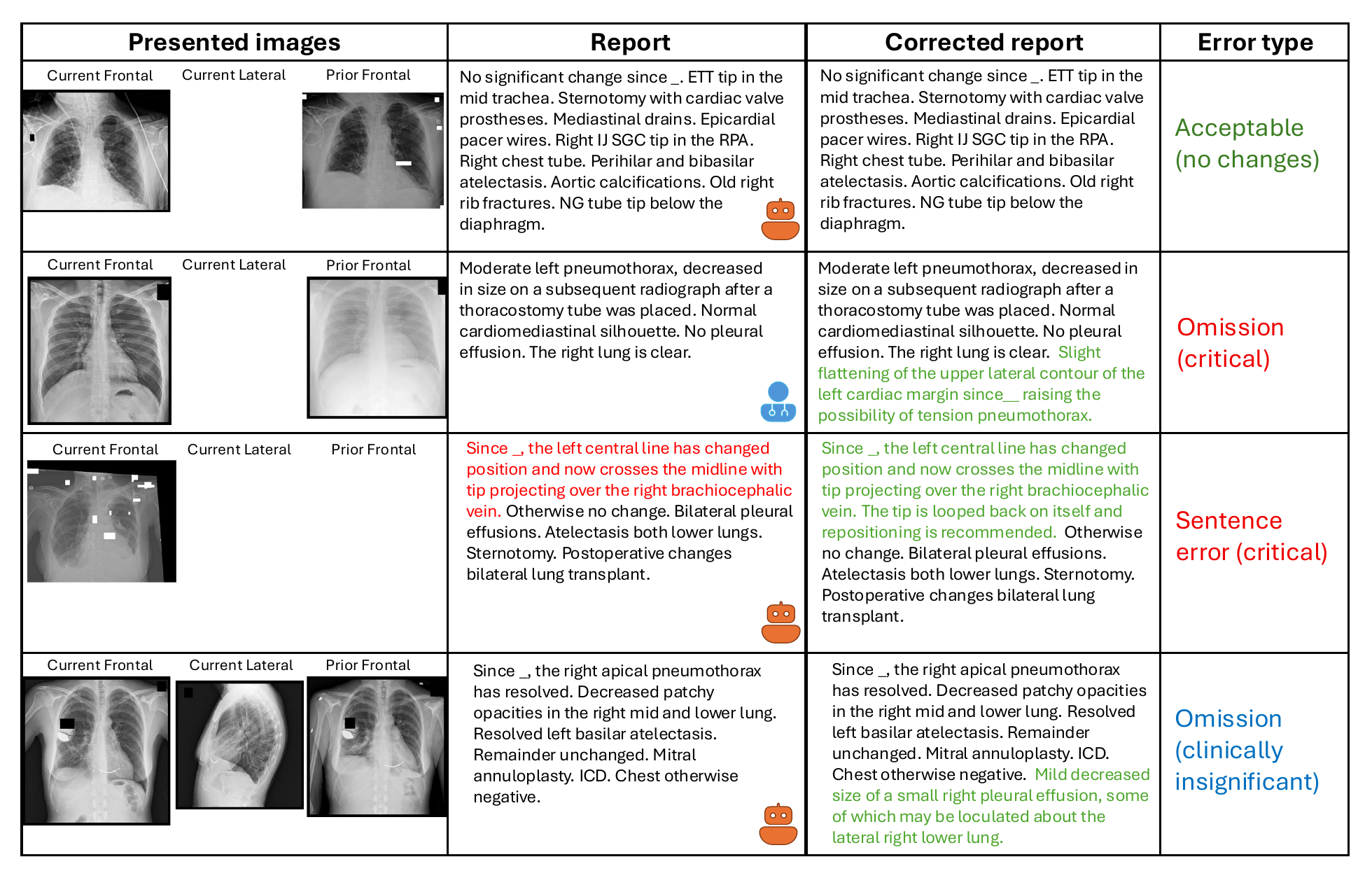}

    \caption{Qualitative examples of errors identified by radiologists. Qualitative examples of original and \mairax generated reports with radiologist-identified errors from the user evaluation study. The source is shown using the orange robot icon (AI-generated) and blue doctor icon (original report). Red text indicates incorrect text; green text indicates text which has been added/corrected in the updated report. Extended qualitative examples are shown in Supplementary Figure S5. Error types (right-most column) are colored using green (no errors), red (clinically significant error(s)), blue (only clinically insignificant errors).}

    \label{fig:qualitative1}

\end{sidewaysfigure}

Furthermore, we categorize all errors identified in the user evaluation study into those related to pathological findings and those related to \acp{LT}, using GPT-4o-based classification. Overall, 63.4\% of errors were pathological and 34.2\% were \ac{LT}-related. When stratified by source, original reports contained 58.4\% pathological and 38.4\% \ac{LT} errors, compared to 67.2\% pathological and 31.1\% \ac{LT} errors in AI-generated reports, representing a significant shift toward pathological errors in AI outputs ($\chi^2 = 5.36$, $p = 0.021$, $\text{degrees of freedom} = 1$). We further analyzed error severity. Among critical errors, 53.9\% were pathological and 43.4\% were \ac{LT}-related, whereas non-critical errors were more heavily skewed toward pathology (65.5\% vs.\ 32.2\% \ac{LT}), indicating a relatively higher contribution of \ac{LT} issues among clinically significant errors ($\chi^2 = 6.68$, $p = 0.010$, $\text{degrees of freedom} = 1$).

A fine-grained analysis of error categories revealed that the most common error types overall were related to other non-\ac{LT} devices, opacity, atelectasis, and pleural effusion. Among critical errors, clinically urgent findings such as opacity and pneumothorax were more prominent, alongside \ac{LT} positioning errors. The distribution of the top error categories for all errors and for critical errors is summarized in \Cref{tab:error_categories_by_source}.

We present extended results of the user evaluation study in Supplementary Section 4.8, with stratification based on radiologists' experience and patient demographics. Lastly, we computed quantitative metrics for both original and AI-generated reports using the modified report from the radiologist evaluators as the reference, for both Target Set and \ac{LT} Set, and report metrics in Supplementary Table S21. 
Lexical and clinical efficacy metrics are high and similar between original and AI-generated reports across both cohorts.
For the \ac{LT}-specific metrics, 
\ac{LT} Set \fone values are greater than $70$ for type, change, overall placement, and counts, while AI reports show lower scores on \ac{LT} incorrect placement.
A detailed analysis of the errors identified by three reviewers per study highlights inter-rater variability among radiologists for the same study. We present these findings, along with qualitative examples, in \Cref{sec:inter_rater}.

\subsection{Analysis of errors reported in the user study highlights high inter-rater variability among radiologists}
\label{sec:inter_rater}
First, we computed the inter-rater agreement on the report scores assigned by radiologists (score 1 for critical/clinically significant errors, 2 for clinically insignificant errors and 3 for acceptable reports). We found an average Kendall's concordance~\cite{kendall90rank} of $W$ = 0.44, with a slightly higher agreement on AI-generated reports than on originals (see Supplementary Table S22 and Supplementary Section 4.8 for details of this analysis and \Cref{sec:user_eval_setup} for the user evaluation study setup). This value indicates a moderate agreement among radiologists on the assigned report scores.

We then computed the inter-rater variability for flagged errors at the sentence level, including both sentence errors and their suggested corrections, as well as omissions. As illustrated in \Cref{fig:error_agreement}, there is substantial disagreement among the radiologists in deciding which sentences require changes and which reports contain omissions. Among all sentence errors or omissions, over $80\%$ were identified by only one of the three reviewers. This pattern was observed in both AI-generated and original reports, with a slightly higher consensus on corrections in the AI-generated reports (multiple reviewers agreed on corrections in 14.6\% of original reports versus 20.0\% of AI-generated reports).
If errors were determined by majority vote, the absolute proportions of reports with critical errors and reports requiring no changes (\Cref{fig:comparison_plot,fig:omissions_sentence_errors}) would appear substantially improved, while the relative gap between \mairax and the original reports would remain approximately unchanged.

Qualitative examples of inter-rater variability are provided in \Cref{tab:inter_rater}. It can be noted that critical and non-critical errors are flagged by radiologists in the original reports as well as in AI-generated reports. For the quantitative analysis illustrated in \Cref{fig:error_agreement}, we considered the reviewers in agreement if they had modified the same sentence or corrected an omission in the same report. The reviewers might still disagree on the specific correction required, as was often observed in our qualitative analysis. 
For instance, we observe a case (example 2) where all three radiologists disagreed on the errors and assigned different scores accordingly, i.e., a clinically insignificant sentence error, a clinically significant omission, and no changes, respectively. Even for the same error type and score (example 3, example 5), they may disagree on the corrections, i.e., different \ac{CVC} tip locations in example 3, and different omissions in example 5. However, we also find agreement among radiologists on scores and corrections (example 1, example 6), i.e., two radiologists agreed on no changes in example 1 and two radiologists agreed on the sentence error and corresponding correction in example 6.

Given the high inter-observer variability in error classification, we conducted an additional consensus analysis to determine the rate of critical errors more precisely. The three most senior radiologists reviewed all the cases that had been flagged as critical by at least one radiologist in the initial review and reclassified these errors based on majority consensus. Using this consensus approach, the proportion of AI-generated reports free from critical errors increased from 95.4$\%$ to 96.9$\%$, while the original reports free from critical errors increased from 97.0$\%$ to 98.8$\%$. These results were found to be equivalent across the two cohorts. Further examination of the confirmed critical errors revealed that the errors related to pathological findings were the most frequent type of critical errors in both AI-generated and original reports. 

\section{Discussion}\label{sec12}

Automatic generation of high-quality narrative-style reports from radiology images could lead to significant improvements in clinical workflows.
However, the successful implementation of clinical AI-driven radiology reporting systems involves addressing several challenges. These include
accurate interpretation of images, generation of linguistically cohesive and clinically relevant reports, and integration of contextual information such as patient history and prior imaging studies when available. They are further complicated by the need for precise and reliable descriptions of lines (catheters) and tubes in \acp{CXR}, specifically in high-volume patient settings such as \acp{ICU} and emergency departments. Leveraging \mairatwo~\cite{bannur2024maira2groundedradiologyreport} -- a state-of-the-art \ac{MLLM} for \ac{CXR} findings generation -- as the base model architecture, \mairax was developed by curating a large-scale, multi-site clinical dataset, fine-tuning the vision encoder and LLM, adjusting hyperparameters and LLM prompts, and designing \ac{LT} performance measures for model optimization, to effectively describe both pathological findings and \acp{LT} in \ac{CXR} studies.
\mairax is a \ac{CXR}-specialized report generation model that not only generates strong draft reports with respect to lexical coherence and clinical quality (achieving superior results on \mimiccxr and \mayodataset for these metrics compared to prior works), but also adequately describes the \ac{LT} information in \ac{CXR} images, as demonstrated by the \ac{LT} metrics of the \ltmetric framework (achieving 10 pp  or more improvements over the baseline on \ac{LT} types and incorrect placements, as well as improvements on longitudinal changes, placements and counts on holdout datasets of \mayodataset).

Insights from the user evaluation study are important for understanding the practical implications of deploying \mairax as an \ac{AI}-assisted reporting tool in clinical settings. The results affirm \mairax's strong performance as a radiology reporting assistant, with a 0.3 pp difference in the proportion of error-free sentences between original and AI-generated reports.  The gap for reports with critical errors was 1.6 pp between original and AI-generated texts (statistically significant, $p = 0.006$), and 5.3 pp for reports deemed acceptable as-is.
While these gaps are narrow, they remain in favor of original reports, and \mairax is positioned as a reporting assistant rather than a replacement for the radiologist. Consistent with this positioning, \mairax is intended to generate draft reports for radiologist review; radiologists are expected to independently verify every finding including line and tube placements before signing the final report. 
Nonetheless, the drafts were of high quality: 79.3\% of \mairax draft reports required no changes, and across the full study the mean number of flagged errors per report was only 0.29, with 0.06 critical errors per report on average. After consensus review, only 3.1\% of AI-generated reports had critical errors. We hypothesize that \mairax could enhance reporting efficiency in routine practice, but any such claim warrants prospective assessment; this retrospective study did not measure whether reviewing a draft is faster or safer than dictating a report.

While recent evaluation studies have emerged for various \ac{AI} radiology report generation models~\cite{huang2025jama,Tanno2025,hong2025radiologist},
none have 
presented a retrospective user evaluation of a state-of-the-art \ac{CXR} report generation model like \mairax encompassing both clinical (pathological findings) and \ac{LT}-specific assessments. Moreover, our user evaluation study
evaluates radiology report generation across two clinically motivated cohorts: one reflecting the intended deployment distribution and one with upsampled \ac{LT} cases.
We found only 4.6\% of \mairax reports containing critical errors (vs. 3\% original reports), a notable improvement compared to previously reported critical error rates of other models (e.g.\ 18\% for Flamingo-CXR in \cite{Tanno2025}, although direct numerical comparison is limited by differences in evaluation cohort, reader pool, and unknown original-report error rate in that study).

We also observed that report quality was higher on the Target Set than on the \ac{LT} Set for both original and AI-generated reports, suggesting that studies with more \acp{LT} are inherently more difficult to interpret for both radiologists and the model. 
The user study additionally revealed substantial inter-rater variability at the sentence-level, with over 80\% of flagged errors identified by only one of the three reviewers; the same pattern was observed for both AI-generated and original reports, indicating intrinsic differences in radiologists' reporting preferences and error-flagging thresholds rather than a deficiency specific to AI content. The disagreement is consistent with known variability in radiologists' reporting styles and suggests that the model's flagged errors and omissions may be distinct from, or complementary to, those identified by the radiologists. Accounting for this variability (e.g., via majority-vote ground truth) could yield apparently improved \mairax performance, though the relative gap to original reports would remain similar.
Stratified analysis further revealed that resident radiologists tended to flag more errors as clinically significant compared to their senior counterparts (Supplementary Table S23), possibly reflecting reduced confidence and experience in distinguishing borderline or insignificant abnormalities from clinically significant findings.
Notably, we also found that the original radiology reports in our dataset were imperfect, with 15\% containing at least one error. This factor ultimately limits the model performance and poses a difficult challenge for radiology multimodal models trained on such large-scale clinical datasets.

In recent years, there has been growing interest in using AI for automatic detection and localization of medical \ac{LT} using traditional computer vision approaches~\cite{yi2020computer}. However, none of these methods offer full-text \ac{CXR} report generation capabilities. Most existing work focuses on detecting a specific \ac{LT} type~\cite{lee2018deep,singh2019assessment,kao2015automated} or a subset of \acp{LT}~\cite{rungta2021detection,henderson2021automatic}. In contrast, we identify nine different \ac{LT} types and the corresponding attributes of each type. We present \ltmetric, an evaluation method focusing on the detailed reporting of \acp{LT} by generative AI models. This comprehensive LLM-based structured report metrics suite provides a fine-grained, scalable framework for evaluating AI-generated reports in capturing \ac{LT} information. We believe our proposed methodology provides a rigorous framework  for the fine-grained \ac{LT}-specific evaluation of generative AI models.

\mairax has been trained on a large-scale clinical dataset from Mayo Clinic. With the development and training of \mairax, we noted data-related challenges and the need to carefully curate our datasets before training the multimodal LLM. One significant challenge lies in dealing with the intricacies of real-world longitudinal and paired multimodal medical data, including quality issues stemming from de-identification protocols (e.g., image occlusions), different acquisition strategies (e.g., pre- and post-Epic integration~\cite{chishtie2023use}), and incomplete contextual information (e.g., linked lateral/prior images, report sections). Therefore, we developed an elaborate quality control and data preprocessing pipeline with several steps such as report cleaning and  filtering, view classification and outlier removal (see~\Cref{sec:data_processing} for details). 

Our study has several limitations. First, although \mayodataset includes data from multiple Mayo Clinic sites, all studies originate from a single institution, which limits external generalizability until further validation on data from other health systems is performed. Furthermore, the user evaluation study subset data came from a single site. Any clinical deployment at other Mayo Clinic sites will require additional user evaluation studies.
The dataset is also demographically imbalanced, with a predominance of White patients (91.5\%), which may limit the generalizability of the model to more diverse populations and raises the possibility of algorithmic bias. While AI systems can underperform on underrepresented groups, our stratified analysis of the user evaluation cohort did not demonstrate degraded performance in non-White patients (Supplementary Figure S4); analyses on the larger Mayo dataset validation and test sets also show comparable performance across demographic strata (Supplementary Tables S9, S10, S11, S12, S13). Nevertheless, these findings should be interpreted with caution given the smaller sample size of non-White subgroups, and further validation on more diverse populations is warranted. The dataset also contains artifacts of de-identification (namely, black and white boxes on images) which, in principle, could be used for spurious shortcut learning. While these overlays are applied systematically as part of a standardized de-identification pipeline and the dataset is multi-site and heterogeneous—reducing the likelihood of a consistent shortcut signal—we cannot fully exclude the possibility that the model leverages such pixel-level artifacts. This represents a potential source of bias, and future work should explicitly evaluate robustness to these artifacts (e.g., through artifact removal, masking, or controlled ablations) to better isolate clinically relevant signal. Additionally, inter-radiologist variability was observed in reporting styles and verbosity, reflecting differences in experience and skill-sets, and affecting the consistency and accuracy of the original reports, particularly in the reporting of \ac{LT} tip locations, and minor or negative findings. Furthermore, the lack of gold standard reports, ground truth labels and performance metrics further complicated the model evaluation process, highlighting the need 
for customized metrics for model optimization and evaluation, particularly for \ac{LT} reporting. 
Our \ltmetric framework relies on a single LLM (GPT-4o) and was not subjected to run-to-run repeatability testing across multiple sampling seeds; agreement was assessed against radiologist-validated structured extractions on two independent test sets (115 Mayo Clinic studies and 100 \mimiccxr studies), but repeatability under stochastic decoding remains uncharacterized. 

Our study design also does not address two factors that will determine clinical benefit in deployment. We did not record reporting or reading times, so the effect of draft review on radiologist efficiency is not established for \mairax; prior evaluation of AI-assisted reporting has reported reduced reporting time~\cite{huang2025jama}, but this must be confirmed prospectively. We also did not assess automation bias among radiologists. Radiologists reviewing a draft known to be \ac{AI}-generated may prematurely approve a report without identifying omitted findings. Both reporting time and automation bias should be studied prospectively.

Lastly, we noted that \mairax exhibited limited performance in detecting incorrectly placed \acp{LT}, likely due to their low prevalence in the dataset (7.7\% of all \acp{LT}) and the inherent inter-rater variability of malposition adjudication. Although the model represents a meaningful absolute improvement over the baseline on this task, the residual gap is clinically consequential: malpositioned \acp{LT} are precisely among findings most likely to affect immediate patient management, and radiologists using \mairax drafts should treat every reported \ac{LT} placement as requiring independent verification on the image. We intend to address this limitation by incorporating additional data with misplaced \acp{LT} following the deployment of \mairax at Mayo Clinic and iteratively enhancing its performance. Future model iterations could also benefit from incorporating reinforcement learning with verifiable rewards and explicit reasoning traces, an approach recently demonstrated for CXR interpretation~\cite{myronenko2025nvreason}, which may further enhance both performance and clinical auditability.

In summary, by improving longitudinal \ac{CXR} report generation for both clinical findings and \acp{LT}, we demonstrate that
\mairax has the potential to serve as a radiologist’s AI assistant for \ac{CXR} draft reporting. By producing reports where 97.4\% of the generated sentences are error-free (vs. 97.7\% in original reports) and 95.4\% of reports do not have any critical errors (vs. 97.0\% original reports), \mairax marks a significant step forward in enhancing the clinical applicability of AI-assisted radiology tools, particularly in high-volume inpatient or \ac{ICU} settings. Given the promising results from our retrospective user-centric evaluation study at Mayo Clinic, we plan to deploy the \mairax model at Mayo Clinic and evaluate its performance 
prospectively, including its effect on reporting time and accuracy, to establish whether it can support streamlined clinical workflows.

\clearpage

\section{Methods}\label{sec11}
\textbf{Ethics declaration} This study involves analysis of de-identified data via the Mayo Clinic Platform\_Discover. In accordance with the Code of Federal Regulations, 45 CFR 46.102, the noted activity does not require IRB review or informed consent.

\subsection{Dataset Details}
\label{sec:dataset_details}

We use a large-scale, multi-site, longitudinal internal dataset of 2,427,411 \ac{CXR} studies comprising 4,123,776 images provided by Mayo Clinic Platform acquired from 744,860 Mayo Clinic patients from Mayo Clinic's clinical archives between 2007 and 2023. We call this dataset \mayodataset. \mayodataset is an internal dataset that is not publicly available; Mayo Clinic Platform can reproduce the dataset for approved external requests under a data-use agreement, as described in the Data Availability statement. Each study contains longitudinal information including current frontal image, current lateral image, prior frontal image, prior reports, and clinical context such as \indication and \comparison sections of reports. Of all studies, 70.0\% are associated with priors and 23.7\% (574,494) have at least one line or tube. The dataset consists of 70.0\% inpatient and 26.3\% outpatient studies. To ensure patient privacy, the reports were de-identified using a de-identification protocol~\cite{murugadoss2021building}. See Supplementary Section 2 for the image deidentification protocol. The EHR-derived demographic distributions for patient age, sex, race, and technical details such as ordering department, year of acquisition, and scanner manufacturers are shown in Supplementary Table S2. We report details for the top 10 of the 126 different ordering departments. The \ac{CXR} images were acquired by scanners from 19 different manufacturers.

A total of 1,184,799 \ac{LT} instances were found in \mayodataset (average of 2.06 \acp{LT}/report in reports with at least one \ac{LT}, and 0.49 \acp{LT}/report overall) after extracting the structured report from the full dataset (see \Cref{sec:quantitative_metrics} for details of \ac{LT} structured reports).  There are 12 \ac{LT} types and 67 \ac{LT} tip locations that were mapped to their corresponding placement type (Supplementary Table S1). 
These 12 types correspond to the nine principal \ac{LT} categories listed in the Introduction, with the \ac{CVC} category further subdivided into Internal Jugular, Subclavian/Port-a-Cath, Unspecified, and Femoral subtypes for the structured-report schema.
``N/A" represents a field not explicitly specified in the report. The distribution of \ac{LT} types, longitudinal change, side, and placement is shown in Supplementary Table S3. Incorrectly placed \acp{LT} account for 7.7\% of all the \acp{LT} in the reports.

\subsection{Data Processing}
\label{sec:data_processing}

\subsubsection{Image Preprocessing}

We first converted chest X-ray images from DICOM to PNG, remapping the minimum and maximum intensities of each image to an 8-bit range of $[0, 2^8)$.
We resized the short side of each image to 518 pixels (as required by our encoder, see \cref{sec: vision_encoder}), maintaining the original aspect ratio.
To prevent aliasing under the downsampling transformation, prior to resampling we applied an in-plane Gaussian prefilter with variance $\sigma^2_P = (k^2 - 1) (2\sqrt{2 \log 2})^{-2}$, where $k$ is the downsampling factor, following the scale-factor point-spread-function matching scheme of~\citet{cardoso2015psf}.
Resampling was performed with a cubic B-spline interpolant~\cite{unser1991,thevenaz2000}, which provides a favorable trade-off between approximation order and computational cost for medical image resampling~\cite{MEIJERING2001111}.
Both operations were implemented using SimpleITK~\cite{Lowekamp2013-tp}.
To ensure patient confidentiality, as part of the de-identification protocol~\cite{murugadoss2021building}, black or white boxes were automatically overlaid on the images to obscure identifying text information.

We found that the raw image dataset contained outliers such as noise images, all black or white images, non-chest X-ray images, and images with very high or low outlier intensity values. Fastdup~\cite{noauthor_visual_nodate} v2.16
was used to detect and remove the outlier images. We reduced the image resolution by a factor of four to speed up processing by Fastdup. We ran Fastdup on all \ac{CXR} images and removed 215,767 outlier images from the dataset (Supplementary Figure S1). 

We found that the DICOM metadata related to view position were incomplete as it was not specified for the majority of the images, hence, we trained a model to classify frontal vs.\ lateral X-rays. The model consists of a pre-trained \raddino encoder~\cite{perez-garcia_exploring_2025} and a linear classification head.
We used the images with existing view DICOM metadata (`View Position' DICOM element) as the training set for the classification model (e.g.\ AP or PA for frontals, LATERAL for laterals). The model achieved an accuracy of 99.4\% on a 20\% held-out validation split by patient. Qualitative inspection of the classification results through image montages showed most frontal views being classified correctly, with slightly more false positives in lateral views specifically for cropped frontal images (e.g.\ frontals showing partial lungs). We used the trained view classifier checkpoint to classify the unlabeled images as frontal or lateral. Using this method, the images were divided into frontal (62\%) and lateral (38\%).

\subsubsection{Report Cleaning and Preprocessing}

The \ac{CXR} reports were preprocessed in two parts: ones from before and after a transition to storing \acp{EHR} using the Epic system~\cite{chishtie2023use}. Reports from post-Epic were formatted as a single string with various headers used to identify the report sections, i.e., variations on “IMPRESSION”, “EXAM TYPE”, “REASON FOR EXAM”, “COMPARISON”, and “FINDINGS”. GPT-4o~\cite{openai2024gpt4ocard} was used to convert each report into a JSON format with fields for each report section as well as clean the content of each section, removing irrelevant artifacts (e.g. text strings unrelated to the X-ray image or the patient's condition such as automatic disclaimers appended to the report, or gibberish strings introduced by conversion of reports to plain text) and duplications, and re-writing content for de-identification such that it is not possible to predict from the context such as electronic signatures, statements of results being discussed with another radiologist, and specific times and dates. A similar process was applied to the reports pre-Epic, however, various challenges included section headers often not being present or only present for some sections, edited reports were appended, and the same information being described multiple times. As such, the GPT-4o prompt included descriptions of how to identify each section and how to clean up inconsistencies and duplicate information. The report cleaning prompt is presented in Supplementary Section 3.1. Prompts were refined iteratively through manual review of processed reports, checking for inclusion of non-clinical information, omissions of relevant clinical content, and hallucinations. Due to inconsistencies in the report section names that could correspond to standard ``TECHNIQUE" (e.g. ``EXAMINATION", ``EXAM", ``PROCEDURE", ``STUDY", ``EXAM TYPE", etc.) and negligible additional information (often just specifying whether a lateral exam is available or not), the ``TECHNIQUE" section was assigned ``N/A" in the \mairax inputs.

During report quality checks, it was observed that 9\% (222,582) of the report findings contained four or fewer words, and on qualitative inspection, these were found to represent normal reports with negative findings. Upon radiologists' suggestion, such normal short reports were replaced by the following standard template text report: ``\texttt{The lungs are clear. Normal cardiomediastinal silhouette. No pneumothorax or pleural effusion.}" Also, short reports without any findings information such as ``\texttt{stable exam}" and ``\texttt{no change}" were filtered out from the dataset. This step removed 107,890 of 2,536,452 studies (4.3\%), of which 99,583 (92.3\%) had no linked \ac{EHR} demographic data; every study lacking linked demographic data had empty report text and was removed here. Among the 2,436,869 studies with linked demographics, the removal rate was 0.341\%, and associations with demographic and facility variables were negligible in magnitude despite reaching statistical significance at this sample size (Cram\'er's $V = 0.006$ for race, $0.007$ for sex, $0.014$ for age, and $0.018$ for facility). Removal rates were lower for four of the five named non-White racial categories than for White patients (0.222\% for Black or African American, 0.184\% for Asian, 0.150\% for Asian (Far East), 0.224\% for Asian (Indian Subcontinent), and 0.377\% for Native American and Pacific Islander, against 0.344\% for White patients), and the White proportion of the dataset was 91.485\% before this step and 91.482\% after it. By age, removal rates were highest in the youngest patients (0.631\% for ages 18--19 and 0.595\% for ages 20--29) and lowest in the seventh and eighth decades (0.285\% for ages 70--79 and 0.280\% for ages 80--89). The highest removal rate of any racial category was for records with race unspecified (0.436\%), and, across all variables examined, the strongest association was with ordering department (Cram\'er's $V = 0.061$; 1.252\% for General Practice and 1.119\% for General Surgery, against 0.114\% for Cardiovascular Diseases), consistent with brief interval-check reports arising disproportionately in routine outpatient and postoperative follow-up.

Finally, upon manual inspection, it was found that the \findings and \impression sections can contain complementary information in the \mayodataset dataset, where the information in the \impression section should have ideally been included in the \findings section. GPT-4o was used to append the additional information from the \impression section to the \findings section of the reports. The GPT-4o prompt is presented in Supplementary Section 3.2.

\subsubsection{Paired Dataset Creation}

After the images and reports were processed, we combined them into a unified dataset of images and report pairs, where we merged the metadata, report text, and multiple views for each \ac{CXR} study. For each frontal image, the corresponding lateral and prior frontal images were linked when available. In many cases, multiple images were acquired during the same clinical visit. To ensure dataset consistency, we applied a de-duplication step that retained only one \ac{CXR} image per type, i.e., a frontal, a lateral, and a prior frontal image per study. This selection was guided by the DICOM metadata, specifically Image Type, Acquisition Date, and Acquisition Time. When original images (\texttt{ImageType = ORIGINAL}) were present, the most recent by timestamp was retained; otherwise, the latest derived image (\texttt{ImageType = DERIVED}) was selected. Images lacking acquisition timestamp data were excluded. An overview of dataset size after various preprocessing steps is shown in Supplementary Figure S1.

\subsection{Model Development}

\subsubsection{RAD-DINO-X Vision Encoder}
\label{sec: vision_encoder}

\raddino~\cite{perez-garcia_exploring_2025} is a self-supervised image-only pre-training approach for \acp{CXR}, based on the DINOv2 \ac{SSL} method~\cite{oquab2023dinov2}. The publicly available checkpoint of \raddino vision encoder~\cite{perez-garcia_exploring_2025}
was trained on approximately 834k \ac{CXR} images sourced from public datasets with frontal and lateral views, with adjustments of the DINOv2 augmentation and training strategy for suitability to \acp{CXR}. \raddino uses an 87M-parameter ViT-Base (ViT-B) backbone and takes images of size 518$\times$518. \raddino uses a patch size of 14$\times$14, leading to a sequence of 37$\times$37 = 1369 visual tokens from each image (we discard the CLS token). At the time of its release, \raddino outperformed image-only and image--text contrastively trained image encoders across multiple \ac{CXR} tasks such as findings classification, image segmentation, and report generation. As a result, this was the choice of the image encoder in \mairatwo~\cite{bannur2024maira2groundedradiologyreport}, and we also adopted the approach for our vision encoder pre-training.

Similar to \mairatwo~\cite{bannur2024maira2groundedradiologyreport}, we pretrain a \raddino vision encoder to use as the frozen encoder in \mairax (\Cref{fig:maira_x_overview}(c)). We call this version of the vision encoder ``R\textsc{ad}-DINO-X". \raddinox has been continually pre-trained starting from the publicly available checkpoint using frontal and lateral \ac{CXR} images from the training split of the \mayodataset dataset. 

\subsubsection{\mairax Multimodal LLM}

\mairax is a \acf{MLLM} built on the \mairatwo~\cite{bannur2024maira2groundedradiologyreport} architecture. The \mairax input design follows prior work in AI-based CXR report generation, which has shown that augmenting the current frontal image with additional context improves AI-generated report quality. Specifically, incorporating the lateral view has been shown to improve automated report generation in AI models~\cite{bannur2024maira2groundedradiologyreport, Liu2025}; providing the \indication section has been shown to improve generated-report quality by allowing the model to selectively report findings relevant to the clinical question~\cite{hyland2024maira1specialisedlargemultimodal,bannur2024maira2groundedradiologyreport}; and providing the prior frontal image, prior report, and \comparison section has been shown to enable accurate description of longitudinal change and to reduce hallucinated references to non-existent priors~\cite{bannur2024maira2groundedradiologyreport}. \mairatwo ~\cite{bannur2024maira2groundedradiologyreport} reports systematic ablations of each of these inputs and demonstrates that their inclusion improves clinical accuracy. Based on this prior AI-model evidence, \mairax adopts the same multimodal input structure: current frontal, current lateral, and prior frontal images, the full prior report, and the \technique, \comparison, and \indication sections.

\Cref{fig:maira_x_overview}(c) shows an overview of the \mairax model architecture. \mairax consists of a \raddinox vision encoder, a randomly initialized four-layer MLP adapter~\cite{bannur2024maira2groundedradiologyreport}, and a 13B-parameter Vicuna v1.5 LLM ~\cite{zheng2023judgingllmasajudgemtbenchchatbot} in a LLaVA-style framework~\cite{liu2023visualinstructiontuning}. \raddinox encodes images into embeddings, the adapter module translates the embeddings into the language representation space, and the image and language tokens are fed to the LLM to generate the \findings section of the radiology report. During \mairax training, the \raddinox vision encoder pretrained on the \mayodataset image-only dataset is kept frozen, and MLP layers and LLM are fine-tuned. Ablations of the vision encoder and of the image and text inputs are reported in Supplementary Table S17 for the validation set and Supplementary Table S18 for the test set. 
We trained \mairax using the training split of the processed dataset containing paired \ac{CXR} images and reports from \mayodataset, as detailed in \Cref{sec:data_processing}. 

We conducted experiments with various LLMs, including Phi-3.5, Llama-2 7B, and Vicuna-13B, and selected Vicuna-13B v1.5 as the LLM for \mairax due to its superior quantitative performance compared with the other models, suggesting that it scales more effectively with our large-scale institutional training dataset. This is in contrast to the observation for \mairatwo-13B~\cite{bannur2024maira2groundedradiologyreport} that did not show significant improvements over the \mairatwo-7B. 

Moreover, the input images were resized instead of being cropped from the center like in \mairatwo, ensuring that any \ac{LT}-related visual information in the \ac{CXR} (e.g. origin or tip locations) is preserved. 

Additionally, we refined the LLM prompt for report generation to explicitly include \ac{LT}-specific information in the \ac{AI}-generated reports, where the exact prompt is shown in Supplementary Section 3.3. To address the low prevalence of incorrectly placed \acp{LT} in the original dataset while ensuring their accurate reporting in the draft reports, we oversampled the subset of samples with incorrectly placed \acp{LT} by a factor of two in the training set. This strategy resulted in improvements in our evaluation metrics, particularly in the scores related to incorrect \ac{LT} placements, compared to the original training dataset. 

Lastly, we carefully selected and optimized the training hyperparameters based on the training and evaluation metrics, including \ac{LT}-specific metrics, for the splits of the large-scale institutional dataset (see \Cref{sec:exp_setup} for details of training hyperparameters). Hence, the \mairax model enhances the \mairatwo framework for improved scalability when training on the large-scale clinical dataset from  Mayo Clinic. Our refinements also ensure that \mairax accurately incorporates clinically relevant and \ac{LT}-specific information into the generated draft reports. 

\subsection{Experimental Setup and Implementation Details}
\label{sec:exp_setup}
After performing the data processing steps of \mayodataset, we used 2,427,411 \ac{CXR} studies with paired images and reports for the report generation experiments. We split the dataset by patient into 81/9/10 splits for training (1,964,699), validation (218,517) and testing (244,195), respectively. We verified that the split strategy preserved metadata variable distributions and ensured consistent separation across experiments. Due to compute-intensive evaluation metrics (e.g.\ LLM-based RadFact and \ac{LT} structured reporting metrics), we randomly sampled 40,007 and 40,006 studies from the corresponding splits to create the validation and test sets, respectively; this ensured similar data distributions to the full splits in these subsets. We further created two holdout sets from the full test split, namely, the ``Target Set" (300 studies) with a distribution of \acp{LT} similar to that expected at the time of institutional clinical deployment and the ``\ac{LT} Set" (300 studies) with a more upsampled distribution of the nine different \ac{LT} types for the quantitative evaluation, and also used these for radiologists' user evaluation (more details of Target Set and \ac{LT} Set are in~\Cref{sec:user_eval_setup}).

For pre-training \raddinox, we used four nodes of eight NVIDIA H100 GPUs per node. We used a batch size of 1280 (40 images per GPU) and continually trained the encoder starting from the public \raddino checkpoint for the equivalent of 100 epochs (from the definition of an epoch in DinoV2~\cite{oquab2023dinov2}). We kept the same training hyperparameters as in \raddino~\cite{perez-garcia_exploring_2025}. We use this trained checkpoint as the frozen \raddinox encoder weights in \mairax.

For training \mairax, we used two nodes of eight NVIDIA H100 GPUs per node. We used Fully Sharded Data Parallel (FSDP) with full sharding for multi-node training. We trained \mairax with a conventional autoregressive cross-entropy loss. We used a batch size of 128 (8 full studies per GPU) and trained for one epoch. During hyperparameter tuning experiments, we found no further improvements in metrics on the validation set after training for more than one epoch. We used AdamW optimizer, a learning rate of $4 \times 10^{-5}$  with a cosine learning rate scheduler, warmup ratio 0.03 and linear RoPE scaling with a factor of 1.5. \mairax training took 2 days and 18 hours.
For inference, we used a single node of eight NVIDIA H100 GPUs and maximum output length of 800 tokens; all reports in our dataset are less than 800 tokens.

For LLM-based report cleaning, combining \impression and \findings sections, RadFact computation, and extracting the \ac{LT} structured reports from the free-text reports, we used one Microsoft Azure OpenAI GPT-4o-2024-08-06~\cite{openai2024gpt4ocard} endpoint.

We used SimpleITK v2.5.2 for all image preprocessing operations. To build, train and evaluate \mairax, we used Python v3.11.11 with PyTorch v2.7.1, numpy v1.23.5, pandas v2.3.1, SciPy v1.14.1, transformers v4.41.2, tokenizers v0.19.1, langchain v0.2.17, openai v1.55.0.

\subsection{Evaluation Framework}

\subsubsection{User Evaluation Study Setup}
\label{sec:user_eval_setup}

For human evaluation of AI-generated reports, we use the Target Set and \ac{LT} Set. The Target Set is intended to mimic the distribution of images in the target clinical setting (i.e., \acp{ICU}, emergency department, and inpatient settings at Mayo Clinic Rochester Campus). For the \ac{LT} Set, we selected studies containing at least one \ac{LT}, ensuring representation of images with rare and incorrectly placed \ac{LT}s. The selected studies included 300 from the \ac{LT} Set and 300 from the Target Set. Both sets were curated by one data scientist in consultation with a senior radiologist (15 years of post-residency experience). In the Target Set, 238 (79.3\%) of studies have no \ac{LT}, while all 300 studies from the \ac{LT} Set have at least one \ac{LT}. The Target Set reflects real-world prevalence of the model's planned deployment environment (\ac{ICU}, emergency, and in-patient settings), whereas the \ac{LT} Set intentionally over-represents \ac{LT} cases to enable more detailed evaluation of model performance on these clinically important but less frequent findings. All imaging studies used for human evaluation were performed in 2023 at the Mayo Clinic Rochester campus, which is the planned initial clinical deployment site for \mairax; this restriction was deliberate so that the user-study cohort matches the imaging hardware, acquisition protocols, clinical workflow, and reporting conventions of the deployment context as closely as possible. General performance across the broader dataset distribution (2007--2023, all Mayo Clinic sites) is characterized separately by the quantitative evaluation on the full 40K test set (Supplementary Table S2). The distribution of the different \acp{LT} in the two sets is shown in Supplementary Figure S2.

Nine radiologists -- six senior radiologists and three residents -- served as reviewers. All reviewers are affiliated with Mayo Clinic Rochester, the planned deployment site, so that report quality was assessed by radiologists calibrated to the local case mix, clinical workflow, and reporting conventions. Senior radiologists had 5, 6, 8, 14, 15, and 29 years of post-residency experience. All residents were in their third year of radiology residency. Each report (i.e., the AI-generated and original report for each image) was evaluated by two senior and one resident radiologist. Reader groups were assembled to be balanced by experience: each group contained one of the three most-senior radiologists, one of the three mid-experience radiologists, and one resident. Reviewers were blinded to whether the report was AI-generated or an original report. Reports were reviewed using an in-house DICOM viewer interface. Reviewers were shown the \ac{CXR} series and, if applicable, the prior \ac{CXR} series. They were also shown the relevant report, split into individual sentences.

Reviewers were instructed to identify errors in reports, which could be corrected by either 1) modifying or deleting an existing sentence or 2) adding an entire sentence which was omitted from the report. Each error was rated as either 1 (clinically significant) or 2 (clinically insignificant). Clinically significant (or critical) errors are defined as errors in findings that need to be notified to the clinician and that may lead to safety issues or errors in patient care management. For example, a report which misses an important finding (e.g. ``large pleural effusion") or a misplaced line (e.g. ``azygous vein placement of a CVC") is considered a clinically significant error. Clinically insignificant errors are defined as errors which are not critical and may not directly affect the patient's health but must be corrected for a report to be deemed acceptable. One example is a missed post-operative hiatal hernia. Reviewers were explicitly told to not make stylistic modifications to the report or penalize for stylistic preferences. 

After noting any error, the radiologists give the report a score between 1 and 3. A 1 indicates critical/clinically significant errors are present, a 2 indicates only clinically insignificant errors are present, and a 3 indicates that the report is acceptable as is. Reviewers also have the option to flag an image if it is unreadable due to quality issues or obstructions introduced by the de-identification process.

The sample size for the user evaluation (300 studies per cohort, 600 total) was chosen as a pragmatic design decision rather than based on a formal power calculation. This size enables reasonably precise estimation of report-level outcomes while allowing separate assessment of each of the two cohorts.

Of the 600 studies reviewed, 39 (6.5\%) were excluded from all analyses because at least one of the three radiologist evaluations was missing or invalid: 31 for a missing or invalid score and 8 for other annotation errors, such as an error mark recorded without an associated significance label. Evaluators were contacted to reconcile the affected annotations, and studies that remained unresolved at the study cut-off date were excluded. Because each study contributed both an original and an AI-generated report, a study with an affected evaluation of either report was excluded from both, keeping the two report types paired and preserving exactly three radiologist evaluations of each report; the affected evaluation concerned the original report in 20 studies, the AI-generated report in 18, and both in 1. All user evaluation study results are therefore computed over 561 studies, corresponding to 1,683 evaluations of original reports and 1,683 evaluations of AI-generated reports. Exclusion was not significantly associated with evaluation subset, patient sex, race, or age (chi-square test, all $p > 0.300$; Cram\'er's $V \le 0.090$); see Supplementary Section 4.7 and Supplementary Tables S19, S20.

\subsubsection{Quantitative Evaluation Metrics}
\label{sec:quantitative_metrics}

\paragraph{Lexical quality and clinical efficacy metrics} To assess the lexical quality of AI-generated reports, we employ the ROUGE-L score~\cite{lin2004automatic}. For evaluating the clinical correctness, we utilize two established clinical efficacy (CE) metrics: the CheXpert \fone-scores~\cite{irvin2019chexpert} based on the CheXbert classifier~\cite{smit2020chexbert}, and RadFact~\cite{bannur2024maira2groundedradiologyreport}, 
a \ac{LLM}-based factuality metric. For the RadFact analysis, the \ac{LLM} splits the AI generated and reference reports into ``atomic statements'' and then evaluates whether these statements are logically supported in either direction, giving a measure of hallucinations or omissions in the generated report. 
These metrics enable us to compare the overall clinical quality of generated reports against original ones effectively.

\paragraph{\ltmetric: An evaluation framework for assessing generative AI models for longitudinal \ac{LT} reporting} 

In the clinical settings, it is important for the reporting radiologists to accurately mention essential aspects of \acp{LT}, including their presence, longitudinal changes, tip locations and placements, as seen in the \ac{CXR} images. Models that can do this could be particularly useful for critical patient management within high-throughput environments, where frequent and precise \ac{LT} reporting is crucial. 
Existing evaluation frameworks do not provide fine-grained, \ac{LT}-specific metrics for quantitatively assessing these elements in radiology report generation.
Although the CheXpert classification~\cite{smit2020chexbert} includes ``support devices" as one of its 14 classes, this category encompasses a wide range of lines, tubes, and other electronic devices (e.g., electrodes, defibrillators, plates, screws, etc.) within a single class. Additionally, the RadFact metric~\cite{bannur2024maira2groundedradiologyreport} developed as a factuality measure, does not provide \ac{LT}-specific performance measurements. To overcome the limitations of existing metrics, we propose an LLM-based evaluation framework, ``\ltmetric{}", designed to assess the \ac{LT}-specific performance of \mairax{}.
This quantitatively captures not only a broad range of \ac{LT} categories, but also clinical attributes beyond device presence or absence, such as their tip locations, longitudinal changes,  placements and counts. 


\ltmetric was developed using an \ac{LT}-specific structured reporting scheme, where we first extracted an \ac{LT}-specific structured report from the free-text report, and then compared individual attributes of the structured AI-generated and original reports to compute the respective metrics.
This ensured that the computed metrics capture whether
each \ac{LT} was meticulously described in the draft report, including aspects such as device name, tip location, side, and changes from prior study. The initial step in the metrics development involved the definition of the structured report schema and categorical fields for different \ac{LT} attributes such as type, tip locations and longitudinal change. 
The LLM-based structured report extraction was performed in two stages. In the first stage, the presence or absence of different \ac{LT} types was established. In the second stage, more fine-grained information such as the tip location, longitudinal change, side, and placement of each detected line/tube was determined to generate the final structured report. Tip locations for different tube types were mapped to their respective placement (i.e.\ correct or incorrect) based on radiologists' feedback. Detailed categories of the extracted \ac{LT} type, tip location, side, longitudinal change, and placement are provided in Supplementary Table S1. ``Unclear" is used when the attribute is specified but its value is not clear from the report text. ``N/A" represents an attribute not explicitly specified in the free-text report.

The \ac{LT} structured report extraction using GPT-4o and evaluation process is illustrated in Supplementary Figure S3.  The LLM prompts used for the structured report extraction, namely the type extraction prompt for the first stage, and an example prompt for the CVC type are presented in Supplementary Section 3.4. The development of the LLM prompt involved the following two steps:
\begin{enumerate}
\item Prompt engineering: This step focused on developing and refining the prompt using a developmental set of 100 studies, followed by a qualitative evaluation and radiologists' feedback.
The prompts, and the report preprocessing rules described above, were developed by two research scientists in consultation with three senior radiologists (14, 15, and 29 years of post-residency experience).
\item Prompt testing: In this step, the developed \ac{LLM} prompt was tested on a holdout test set of 115 studies that were manually structured by one research scientist and evaluated by the same three senior radiologists. For the prompt testing stage on the holdout test set, the \fone-scores achieved were 0.94 for \ac{LT} type, 0.91 for \ac{LT} tip location, 0.94 for \ac{LT} side, 0.88 for longitudinal change, and 0.92 for \ac{LT} placement.  
\end{enumerate}

We further performed testing of \ltmetric on 100 reports from \mimiccxr. Specifically, we randomly selected reports with and without lines and tubes (25 reports without lines and tubes, 75 reports with lines and tubes). One senior radiologist (15 years of post-residency experience) manually evaluated and edited each extraction. For this test, the \fone-scores achieved were 0.99 for \ac{LT} type, 1.0 for \ac{LT} tip location, 1.0 for \ac{LT} side, 1.0 for longitudinal change, and 0.97 for \ac{LT} placement. 
The composition of both the internal holdout test set (115 studies) and the external \mimiccxr validation set (100 studies), including the study-level distribution and per-device breakdown of lines and tubes, is summarized in Supplementary Table S16.

To compute the \ac{LT} structured report metrics, we generated structured reports from both original and AI-generated free-text reports and compared their categorical fields. We computed the macro \fone-scores on \ac{LT} type (specifically for PICC, Chest tube, ETT, NGT, CVC, IABP, SGC, Mediastinal Drain, Tracheostomy). For each matched \ac{LT} instance, we also computed macro \fone-scores for longitudinal change and placement. Additionally, we report the average accuracy of \ac{LT} counts in the report, categorized as 0, 1, 2, 3-or-more \acp{LT}. This comprehensive LLM-based evaluation provides a fine-grained framework for assessing AI-generated reports in capturing detailed \ac{LT} information, which is of high importance in the clinical \ac{CXR} reporting scenario.

\clearpage

\backmatter

\section*{Data Availability}
The chest X-ray studies and reports used in this work were provided by Mayo Clinic Platform, which can reproduce the dataset used in this paper for approved requests. The data are fully de-identified, with no direct identifiers and no means of re-identification, in accordance with the U.S. Common Rule and HIPAA Safe Harbor standards, but remain governed by Mayo Clinic Platform data-use agreements and cannot be redistributed publicly by the corresponding authors. Researchers interested in accessing the data may submit a request directly through Mayo Clinic Platform.

\section*{Code Availability}
The code for model architectures, training and evaluation has been developed starting from the open-source LLaVA framework at \url{https://github.com/haotian-liu/LLaVA}. The RadFact metric is open-sourced and available at~\url{https://github.com/microsoft/radfact}.
Public checkpoints used for \mairatwo and \raddino are available at \url{https://huggingface.co/microsoft/maira-2} and \url{https://huggingface.co/microsoft/rad-dino}, respectively. The trained \mairax model weights are currently held as proprietary by Mayo Clinic and are not being released externally at this time. Institution-specific code used for data preprocessing on Mayo Clinic Platform inputs and for result analysis is similarly not being released, as it is tailored to internal data formats and infrastructure; however, the preprocessing pipeline, training procedure, and RAD-LT-EVAL extraction and scoring procedure are described in detail in the Methods to be reimplemented on any independent dataset. We provide the software packages used along with their versions in \Cref{sec:exp_setup} and LLM prompts used in Supplementary Section 3.4.



\section*{Acknowledgments}
Microsoft would like to express gratitude to Matthew Lungren for his valuable expertise and insights, Anja Thieme for her useful suggestions and advice during project conception,  Sophie Ghazal for scoping the project and setting up the collaboration with Mayo Clinic, Hannah Richardson for her assistance with institutional approvals, and Felix Meissen for discussions and feedback on the RadFact metric. Mayo Clinic would like to thank Patrick Duffy, Michael Lucas, Marc Blasi, Nishant Nadkarni, Chris Roering, Dana Swanstrom, Jennifer Flores and Mark Ibrahim for their engineering and project management contributions. We also thank Mayo Clinic Platform for providing the data used in this study.

\section*{Funding Information}
The research was funded by Mayo Clinic and Microsoft. This work is supported in part by the generosity of Stephen A. and Linda L. Odell.

\section*{Author Contributions Statement}

\subsection*{Authors and Affiliations}

\textbf{Microsoft:} H.S., V.S., A.S., M.I., O.M., S.B.T., F.P.G., M.T.W., N.C., M.J.,  S.B., K.B., D.C.C., S.H., J.A.V.\\
\textbf{Mayo Clinic:} M.C.R., A.G.S., K.K.H., V.K.M., A.C., C.C., S.A.S., M.B.N., A.J.G., H.A.O., B.S.E., B.A.S., P.K., A.K.

\subsection*{Author Contributions}
H.S., V.S. and M.C.R. equally contributed to this work. H.S. and V.S. led the \mairax model development and quantitative evaluation. M.C.R. and V.S. led the user evaluation study, including protocol design, data preparation and analysis of the results. H.S., V.S. and M.C.R. led the paper drafting. A.G.S., K.K.H. and A.K. are senior radiologists who provided clinical supervision for the work and evaluated RAD-LT-EVAL. A.S., O.M.,  M.I., H.S., S.B.T., F.P.G. and V.S. performed data preprocessing, infrastructure setup, and label extraction. M.I. performed \raddinox encoder training. V.K.M., A.C., C.C., S.A.S., M.B.N. and A.J.G. are radiologists who supported the user evaluation study. H.A.O. and B.S.E. provided support with the DICOM user interface and setup for the user evaluation study. B.A.S. is a medical physicist, who provided expert consultation on image characteristics and acquisition parameters.
M.T.W. provided clinical supervision to the Microsoft team during the project. N.C. provided technical feedback on the encoder training.
M.J. helped in designing the user evaluation study. S.B., K.B., D.C.C. and S.H. provided support with \mairatwo and RadFact code. P.K., A.K. and J.A.V. are shared senior authors who led the project conceptualization, provided equal supervision, and secured project funding.
A.G.S., K.K.H., A.S., M.I., O.M., S.B.T., F.P.G., M.T.W., N.C., M.J., K.B., D.C.C., S.H., P.K., A.K., J.A.V. provided feedback on the paper draft. 

\section*{Competing Interests Statement}
The authors declare no competing interests.

\section*{Tables}
\begin{table}[htbp]
  \centering
  \renewcommand{\arraystretch}{1.2} 
\resizebox{\linewidth}{!}{ 
     \begin{tabular}{l|l|p{1.2cm}|p{1.2cm}|p{1.2cm}|l|l|l}  
\toprule
\textbf{Metric} & \multicolumn{1}{l|}{\textbf{MedGemma 4B}~\cite{sellergren2025medgemmatechnicalreport}} & \multicolumn{1}{p{1.2cm}|}{\textbf{LLaVA-Rad}~\cite{ZambranoChaves2025}} & \multicolumn{1}{p{1.2cm}|}{\textbf{Med-PaLM M}~\cite{tu2023generalistbiomedicalai}} & \multicolumn{1}{p{1.2cm}|}{\textbf{Libra}~\cite{zhang2025libraleveragingtemporalimages}} & \textbf{\mairatwo}~\cite{bannur2024maira2groundedradiologyreport} & \textbf{\mairax} & \multicolumn{1}{p{1.8cm}}{\textbf{MAIRA-X No FT}} \\
    \midrule
    ROUGE-L & \hfill 13.9 \graybrackets{[13.7, 14.1]}& \hfill 30.6 & \hfill 27.29 & \hfill 36.2 & 38.4 \graybrackets{[37.8, 39.1]} & \textbf{41.3} \graybrackets{[41.0, 41.6]} & 20.0 \graybrackets{[19.8, 20.1]} \\
    \midrule
    CheXpert/macro-\fone-14 & \hfill 35.6 \graybrackets{[34.1, 37.0]} & \hfill 39.5 & \hfill 39.83 & \hfill 40.2 & 42.7 \graybrackets{[40.9, 44.4]} & \textbf{47.2} \graybrackets{[46.5, 47.9]} & 40.2 \graybrackets{[39.5, 40.9]} \\
    CheXpert/micro-\fone-14 & \hfill 51.1 \graybrackets{[49.9, 52.2]} & \hfill 57.3 & \hfill 53.56 & \hfill 55.3 & 58.5 \graybrackets{[57.3, 59.6]} & \textbf{64.1} \graybrackets{[63.6, 64.5]} & 58.8 \graybrackets{[58.4, 59.3]} \\
    CheXpert/macro-\fone-5 & \hfill 44.2 \graybrackets{[42.7, 45.9]} & \hfill 47.7 & \hfill 51.6 & \hfill 52.6 & 51.5 \graybrackets{[49.3, 53.5]} & \textbf{53.2} \graybrackets{[52.3, 54.0]} & 45.1 \graybrackets{[44.3, 45.9]} \\
    CheXpert/micro-\fone-5 & \hfill 52.9 \graybrackets{[51.4, 54.5]} & \hfill 57.4 & \hfill 57.88 & \hfill 58.9 & 58.9 \graybrackets{[57.4, 60.5]} & \textbf{61.8} \graybrackets{[61.1, 62.4]} & 55.6 \graybrackets{[54.9, 56.3]} \\
    \midrule
    RadFact/logical-precision & \hfill - & \hfill - & \hfill - & \hfill - & 52.5 \graybrackets{[51.6, 53.5]} & \textbf{61.0} \graybrackets{[60.6, 61.4]} & 52.9 \graybrackets{[52.4, 53.3]} \\
    RadFact/logical-recall & \hfill - & \hfill - & \hfill - & \hfill - & 48.6 \graybrackets{[47.7, 49.6]} & \textbf{55.1} \graybrackets{[54.7, 55.5]} & 47.1 \graybrackets{[46.7, 47.5]} \\
    \bottomrule
    \end{tabular}%
    }
    \caption{Quantitative results of \mairax compared to prior works in the literature on the \mimiccxr official test split. \mairatwo and \mairax values are mean, 95\% \ac{CI} for $n=500$ bootstrapped samples are shown. \mairax No FT shows MAIRA-X performance without finetuning on the MIMIC-CXR training split. All baseline models were trained on MIMIC-CXR data; MAIRA-X No FT is the only model evaluated without any MIMIC-CXR training. Highest scores for each metric are shown in bold.}
  \label{tab:mairax_mimic}%
\end{table}%

\begin{table*}[t]
\centering

\small
\setlength{\tabcolsep}{6pt}
\begin{tabular}{p{0.38\textwidth} c p{0.38\textwidth} c}
\toprule\multicolumn{4}{c}{\textbf{Original Reports}} \\
\cmidrule(lr){1-2} \cmidrule(lr){3-4}
\textbf{All Errors} & \textbf{(\%)} & \textbf{Critical Errors} & \textbf{(\%)} \\
\midrule
Other device & 7.9 & Opacity & 19.0 \\
Atelectasis & 7.9 & \ac{CVC} positioning & 10.3 \\
Opacity & 7.7 & \ac{ETT} tube positioning & 8.6 \\
Pleural effusion & 7.4 & Other device & 8.6 \\
CVC positioning & 7.1 & \ac{PICC} line placement & 8.6 \\
\ac{ETT} tube positioning & 4.7 & Pneumothorax & 6.9 \\
Pneumothorax & 3.8 & Mass/nodules & 5.2 \\
\ac{PICC} line placement & 3.6 & Atelectasis & 5.2 \\
Postoperative changes & 3.6 & Presence/absence of \ac{LT} & 3.4 \\
Pacemaker/lead & 3.6 & Image visual quality & 3.4 \\
\midrule
\multicolumn{4}{c}{\textbf{AI-Generated Reports}} \\
\cmidrule(lr){1-2} \cmidrule(lr){3-4}
\textbf{All Errors} & \textbf{(\%)} & \textbf{Critical Errors} & \textbf{(\%)} \\
\midrule
Opacity & 10.2 & Pneumothorax & 18.1 \\
Pleural effusion & 9.4 & \ac{CVC} positioning & 14.9 \\
CVC positioning & 7.7 & Opacity & 14.9 \\
Atelectasis & 5.6 & Mass/nodules & 6.4 \\
Pneumothorax & 5.6 & Pleural effusion & 6.4 \\
Cardiomegaly & 4.8 & \ac{NGT} placement & 5.3 \\
Other device & 4.4 & \ac{ETT} tube positioning & 5.3 \\
Mass/nodules & 3.8 & Presence/absence of \ac{LT} & 5.3 \\
Surgical clips & 3.5 & Chest tube placement & 4.3 \\
\ac{ETT} positioning & 3.5 & \ac{SGC} positioning & 3.2 \\
\bottomrule

\end{tabular}
\caption{User evaluation study Top-10 error categories stratified by report source (Original vs. AI-generated) and clinical significance (All Errors vs. Critical Errors), each sorted by frequency within subgroup.}
\label{tab:error_categories_by_source}
\end{table*}

\begin{table}[b]
    \centering
    \begin{tabular}{llrrrrrrr}
        \toprule
        & & & \multicolumn{2}{c}{\textbf{1 Reviewer}} & \multicolumn{2}{c}{\textbf{2 Reviewers}} & \multicolumn{2}{c}{\textbf{3 Reviewers}} \\
        \cmidrule(lr){4-5}\cmidrule(lr){6-7}\cmidrule(lr){8-9}
        \textbf{Set} & \textbf{Report} & \textbf{Total} & $n$ & \% & $n$ & \% & $n$ & \% \\
        \midrule
        \multirow{2}{*}{Both Sets}
            & Original     & 302 & 258 & 85.4 & 35 & 11.6 & 9 & 3.0 \\
            & AI-Generated & 369 & 295 & 79.9 & 65 & 17.6 & 9 & 2.4 \\
        \midrule
        \multirow{2}{*}{Target Set}
            & Original     & 105 & 88  & 83.8 & 12 & 11.4 & 5 & 4.8 \\
            & AI-Generated & 155 & 124 & 80.0 & 28 & 18.1 & 3 & 1.9 \\
        \midrule
        \multirow{2}{*}{L\&T Set}
            & Original     & 197 & 170 & 86.3 & 23 & 11.7 & 4 & 2.0 \\
            & AI-Generated & 214 & 171 & 79.9 & 37 & 17.3 & 6 & 2.8 \\
        \bottomrule
    \end{tabular}
    \caption{Agreement on error corrections between the three reviewing radiologists, by evaluation set and report type. For each set (Both, Target, L\&T), counts and percentages show how many flagged sentences or omissions were identified by exactly 1, 2, or 3 reviewers. Modifications to the same sentence or addition of the same omission were considered agreement; disagreement on the specific correction content may still occur.}
    \label{fig:error_agreement}
\end{table}

\begin{table}[h!]
  \centering
  \normalsize
  \resizebox{\linewidth}{!}{ 
    \begin{tabular}{@{}p{0.5cm}|p{4.5cm}|p{4.5cm}|p{4.5cm}|p{4.5cm}|p{2cm}@{}}
    \toprule
    \textbf{Nr.} & \multicolumn{1}{l|}{\textbf{Unmodified Report}} & \multicolumn{1}{l|}{\textbf{Radiologist 1}} & \multicolumn{1}{l|}{\textbf{Radiologist 2}} & \multicolumn{1}{l|}{\textbf{Radiologist 3}} & \multicolumn{1}{l}{\textbf{Agreement}} \\
    \midrule
    1  & \textbf{Original}: Moderate left pneumothorax, decreased in size on a subsequent radiograph after a thoracostomy tube was placed. Normal cardiomediastinal silhouette. No pleural effusion. The right lung is clear.
   &  Moderate left pneumothorax, decreased in size on a subsequent radiograph after a thoracostomy tube was placed. Normal cardiomediastinal silhouette. No pleural effusion. The right lung is clear. \uline{Slight flattening of the upper lateral contour of the left cardiac margin since\_\_ raising the possibility of tension pneumothorax.} \textbf{Score: 1, Omission, Consensus Critical}
 &  Moderate left pneumothorax, decreased in size on a subsequent radiograph after a thoracostomy tube was placed. Normal cardiomediastinal silhouette. No pleural effusion. The right lung is clear. \textbf{Score: 3, No changes}    &    Moderate left pneumothorax, decreased in size on a subsequent radiograph after a thoracostomy tube was placed. Normal cardiomediastinal silhouette. No pleural effusion. The right lung is clear. \textbf{Score: 3, No changes}    &  Radiologist 1 flags a critical omission, Radiologist 2 and 3 agree on no changes required.  \\
    \midrule
   2 &  \textbf{AI-generated}: Since earlier today, new right IJ CVC with tip directed laterally in the right axillary vein. Recommend repositioning. No pneumothorax. Increased pulmonary vascular congestion and interstitial edema. Remainder unchanged. Bibasilar atelectasis. &  Since earlier today, new right IJ CVC with tip directed laterally in the right \uwave{subclavian vein}. Recommend repositioning. No pneumothorax. Increased pulmonary vascular congestion and interstitial edema. Remainder unchanged. Bibasilar atelectasis. 
    \textbf{Score: 2, Sentence error} &   Since earlier today, new right IJ CVC with tip directed laterally in the right axillary vein. Recommend repositioning. No pneumothorax. Increased pulmonary vascular congestion and interstitial edema. Remainder unchanged. Bibasilar atelectasis. \uline{Right costophrenic angle is outside the field of view. Recommend additional imaging.} \textbf{Score: 1, Omission, No Consensus Critical}  &   Since earlier today, new right IJ CVC with tip directed laterally in the right axillary vein. Recommend repositioning. No pneumothorax. Increased pulmonary vascular congestion and interstitial edema. Remainder unchanged. Bibasilar atelectasis. \textbf{Score: 3, No changes}  & No agreement in type of error and scores. \\
    \midrule
  3 & \textbf{AI-generated}: Since earlier today, new left IJ CVC with tip in the upper SVC. No pneumothorax. No other change. ETT with tip in good position. Right IJ SGC with tip in the MPA. Sternotomy with mediastinal clips, drains and AVR. Left atrial appendage closure clip. AV pacemaker. Left chest tube. Small right pleural effusion with associated atelectasis in the right base.  &  Since earlier today, new left IJ CVC with tip in the \uline{right brachiocephalic vein}. No pneumothorax. No other change. ETT with tip in good position. Right IJ SGC with tip in the MPA. Sternotomy with mediastinal clips, drains and AVR. Left atrial appendage closure clip. AV pacemaker. Left chest tube. Small right pleural effusion with associated atelectasis in the right base. \textbf{Score: 1, Sentence error, No Consensus Critical}  &  Since earlier today, new left IJ CVC with tip \uline{malpositioned and looping into the lower right IJ}. No pneumothorax. No other change. ETT with tip in good position. Right IJ SGC with tip in the MPA. Sternotomy with mediastinal clips, drains and AVR. Left atrial appendage closure clip. AV pacemaker. Left chest tube. Small right pleural effusion with associated atelectasis in the right base. \textbf{Score: 1, Sentence error, Consensus Critical}    &  Since earlier today, new left IJ CVC with tip in the upper SVC. No pneumothorax. No other change. ETT with tip in good position. Right IJ SGC with tip in the MPA. Sternotomy with mediastinal clips, drains and AVR. Left atrial appendage closure clip. AV pacemaker. Left chest tube. Small right pleural effusion with associated atelectasis in the right base. \textbf{Score: 3, No changes}    &  Radiologists 1 and 2 agree in the type of error and scores, but their corrected tip locations are different. \\
    \midrule
  4 & \textbf{Original}: No focal consolidation. No large pleural effusion or discernible pneumothorax. Mild bibasilar atelectasis. Unremarkable cardiac silhouette size.   & \uline{Parenchymal opacity in the medial right lower lung may be due an area of infection/pneumonia or atelectasis.} No large pleural effusion or discernible pneumothorax. Mild bibasilar atelectasis. Unremarkable cardiac silhouette size. \textbf{Score: 1, Sentence error, Consensus Critical} &   \uline{Retrocardiac opacification.} No large pleural effusion or discernible pneumothorax. Mild bibasilar atelectasis. \uline{ Enlarged }cardiac silhouette. \textbf{Score: 1, Sentence error, Consensus Critical}   &  No focal consolidation. No large pleural effusion or discernible pneumothorax. Mild bibasilar atelectasis. Unremarkable cardiac silhouette size.  \textbf{Score: 3, No changes}    &  Radiologists 1 and 2 agree on the critical error and scores. \\
    \midrule
  5 &  \textbf{Original}: Compared with \_. The right IJ CVC has been removed. No focal airspace consolidation. No pleural effusion or pneumothorax. Hyperinflation. Scattered bilateral calcified granulomas. Normal heart size. Sternotomy.   & Compared with \_. The right IJ CVC has been removed. No focal airspace consolidation. No pleural effusion or pneumothorax. Hyperinflation. Scattered bilateral calcified granulomas. Normal heart size. Sternotomy.  \uwave{Healed left proximal humerus fracture.} \textbf{Score: 2, Omission} &   Compared with \_. The right IJ CVC has been removed. No focal airspace consolidation. No pleural effusion or pneumothorax. Hyperinflation. Scattered bilateral calcified granulomas. Normal heart size. Sternotomy.  \uwave{No free air under the diaphragm.} \textbf{Score: 2, Omission}   & Compared with \_. The right IJ CVC has been removed. No focal airspace consolidation. No pleural effusion or pneumothorax. Hyperinflation. Scattered bilateral calcified granulomas. Normal heart size. Sternotomy. \textbf{Score: 3, No changes}   &  Radiologists 1 and 2 report different omissions. Radiologist 3 does not find errors. \\
    \midrule
  6 &  \textbf{AI-generated}: Since \_, the left central line has changed position and now crosses the midline with tip projecting over the right brachiocephalic vein. Otherwise no change. Bilateral pleural effusions. Atelectasis both lower lungs. Sternotomy. Postoperative changes bilateral lung transplant.   & Since \_, the left central line has changed position and now crosses the midline with tip projecting over the right brachiocephalic vein. \uline{The tip is looped back on itself and repositioning is recommended.}  Otherwise no change. Bilateral pleural effusions. Atelectasis both lower lungs. Sternotomy. Postoperative changes bilateral lung transplant. \textbf{Score: 1, Sentence Error, Consensus Critical} &   Since \_, the left central line has changed position, \uline{in which the catheter is looped within the right brachiocephalic vein and tip projecting at the brachiocephalic confluence.} Otherwise no change. \uwave{Small right and moderate left} pleural effusions. Atelectasis both lower lungs. Sternotomy. Postoperative changes bilateral lung transplant. \textbf{Score: 1, Sentence Errors, Consensus Critical}   &  Since \_, the left central line has changed position and now crosses the midline with tip projecting over the right brachiocephalic vein. Otherwise no change. Bilateral pleural effusions. Atelectasis both lower lungs. Sternotomy. Postoperative changes bilateral lung transplant. \textbf{Score: 3, No changes}   &  Radiologists 1 and 2 report same sentence error. Radiologist 2 reports additional sentence error. Radiologist 3 does not find errors.  \\
    \bottomrule
    \end{tabular}%
    }
\caption{Qualitative examples for inter-rater variability in user evaluation. Solid underline: critical errors; wavy underline: clinically insignificant errors. For critical errors, we denote whether they were determined to be critical by consensus of senior radiologists (``Consensus Critical'') or not (``No Consensus Critical''). Score, error type, and consensus criticality are shown in bold.}
\label{tab:inter_rater}%
\end{table}%

\clearpage
\section*{Figure Legends}
\noindent\textbf{Figure 1. MAIRA-X overview.} Overview of \mairax for longitudinal chest X-ray (\ac{CXR}) reporting including clinical findings and lines and tubes (\ac{LT}). (a) Problem definition (inputs and output shown for illustration only) (b) \mayodataset dataset: a large-scale, multi-site, clinical dataset from Mayo Clinic (c) \mairax model architecture (d) Summary of \mairax user evaluation study.

\noindent\textbf{Figure 2. \mairax outperforms \mairatwo.} Comparison of \mairax and \mairatwo on the \mayodataset holdout test set. Numeric values/center lines are median (ROUGE-L, CheXpert, RadFact) and mean (lines and tubes (\ac{LT}) metrics); error bars are 95\% \ac{CI} for $n=500$ bootstrapped samples. Raw bootstrap sample values are shown by small dots. Extended results, including those for other holdout datasets are presented in Supplementary Tables S4 and S5.

\noindent\textbf{Figure 3. Schematic overview of the \mairax user evaluation study.} Radiologists see the original and AI-generated report from each chest X-ray (\ac{CXR}) study. Each report is rated by three radiologists in the user study, then errors are binned into categories using GPT-4o.

\noindent\textbf{Figure 4. Radiologist-assessed error rates in original and MAIRA-X-generated reports.} For original and \mairax-generated reports, proportions of (a) sentences requiring change (b) reports with critical errors and (c) reports with changes needed (i.e. any errors) are shown. Numeric values/center lines are the non-bootstrapped metrics. Error bars indicate 95\% confidence intervals obtained from 1,000 bootstrap resamples of the dataset. Raw bootstrap sample values are shown by small dots.

\noindent\textbf{Figure 5. Omissions and sentence errors in original versus \mairax reports.} Proportions of reports with (a) omissions and (b) sentence errors across original and \mairax-generated reports are shown. Numeric values/center lines are the non-bootstrapped metrics. Error bars indicate 95\% confidence intervals obtained from 1,000 bootstrap resamples of the dataset. Raw bootstrap sample values are shown by small dots.

\noindent\textbf{Figure 6. Qualitative examples of errors identified by radiologists.} Qualitative examples of original and \mairax generated reports with radiologist-identified errors from the user evaluation study. The source is shown using the orange robot icon (AI-generated) and blue doctor icon (original report). Red text indicates incorrect text; green text indicates text which has been added/corrected in the updated report. Extended qualitative examples are shown in Supplementary Figure S5. Error types (right-most column) are colored using green (no errors), red (clinically significant error(s)), blue (only clinically insignificant errors).
\clearpage

\end{document}